\documentclass[letterpaper]{article} 
\usepackage{aaai25}  
\usepackage{times}  
\usepackage{helvet}  
\usepackage{courier}  
\usepackage[hyphens]{url}  
\usepackage{graphicx} 
\usepackage{multicol}
\usepackage{multirow}
\usepackage{amsmath}
\usepackage{amssymb}
\usepackage{dsfont}
\usepackage{pifont}  
\usepackage{booktabs}
\usepackage{xcolor}
\usepackage{makecell}

\newcommand{\ours}{UniDet3D}
\newcommand{\cmark}{\ding{51}}

\newcommand{\reimpl}[1]{\textcolor{gray}{#1}}  
\definecolor{blue}{rgb}{0.21, 0.49, 0.74}
\usepackage{natbib}  
\usepackage{caption} 
\usepackage{algorithm}
\usepackage{algorithmic}

\usepackage{newfloat}
\usepackage{listings}
\DeclareCaptionStyle{ruled}{labelfont=normalfont,labelsep=colon,strut=off} 
\floatstyle{ruled}
\newfloat{listing}{tb}{lst}{}
\floatname{listing}{Listing}
\title{UniDet3D: Multi-dataset Indoor 3D Object Detection}
\author{
    Maksim Kolodiazhnyi\textsuperscript{\rm 1},
    Anna Vorontsova,
    Matvey Skripkin\textsuperscript{\rm 1},
    Danila Rukhovich,
    Anton Konushin\textsuperscript{\rm 1}
}
\affiliations{
    \textsuperscript{\rm 1}Artificial Intelligence Research Institute

    \{kolodiazhnyi, skripkin, konushin\}@airi.net
}

\nocopyright

\begin{document}

\maketitle

\begin{abstract}

Growing customer demand for smart solutions in robotics and augmented reality has attracted considerable attention to 3D object detection from point clouds. Yet, existing indoor datasets taken individually are too small and insufficiently diverse to train a powerful and general 3D object detection model. In the meantime, more general approaches utilizing foundation models are still inferior in quality to those based on supervised training for a specific task.
In this work, we propose \ours{}, a simple yet effective 3D object detection model, which is trained on a mixture of indoor datasets and is capable of working in various indoor environments.
By unifying different label spaces, \ours{} enables learning a strong representation across multiple datasets through a supervised joint training scheme. The proposed network architecture is built upon a vanilla transformer encoder, making it easy to run, customize and extend the prediction pipeline for practical use. Extensive experiments demonstrate that \ours{} obtains significant gains over existing 3D object detection methods in 6 indoor benchmarks: ScanNet (+1.1 mAP\textsubscript{50}), ARKitScenes (+19.4 mAP\textsubscript{25}), S3DIS (+9.1 mAP\textsubscript{50}), MultiScan (+9.3 mAP\textsubscript{50}), 3RScan (+3.2 mAP\textsubscript{50}), and ScanNet++ (+2.7 mAP\textsubscript{50}). Code is available at \url{https://github.com/filapro/unidet3d}.

\end{abstract}

%

\section{Introduction}

3D object detection from point clouds aims at simultaneous localization and recognition of 3D objects given a 3D point set. As a core technique for 3D scene understanding, it is widely applied in robotics, AR, and 3D scanning.

Due to major variations in scale and visual appearance of indoor scenes, complemented with different selections and placement of objects, indoor 3D data tends to be complex and diverse. Besides, captured by various sensors ranging from Kinect to generic smartphone cameras, indoor data is inconsistent regarding point cloud density and scene coverage. This leads to a domain gap between different datasets.

Indoor benchmarks contain at most thousands of scenes, e.g., the popular ScanNet~\cite{dai2017scannet} has 1513 scenes, a more recent ARKitScenes~\cite{baruch2021arkitscenes} has 5042 scenes, while S3DIS~\cite{armeni2016s3dis}, ScanNet++~\cite{yeshwanthliu2023scannetpp} are the order of magnitude smaller. None of the datasets contains data of sufficient diversity and volume to train a general model which can be transferred between datasets without severe loss of quality.

Applying a 3D scene understanding model outside the single training domain is possible to a certain extent with visual-language models, that encapsulate the fundamental knowledge about the world via establishing relations between imagery and textual data. 

However, visual-language models imply open-vocabulary problem formulation rather than limiting label spaces to predefined categories. In 3D scene understanding, visual-language models are used to precompute 2D image features, which are lifted to 3D space. Still, these 2D-to-3D approaches in 3D instance segmentation~\cite{nguyen2024open3dis, takmaz2024openmask3d} and 3D object detection~\cite{lu2023ov-3det, zhang2024fm-ov3d} are inferior to supervised baselines~\cite{schult2023mask3d, misra20213detr}. In the meantime, the size of existing real-world indoor datasets is currently insufficient for training visual-language models that can provide high-quality 3D features directly~\cite{jia2024sceneverse}. 

\begin{figure}[t!]
    \centering
        \includegraphics[width=1.\linewidth]{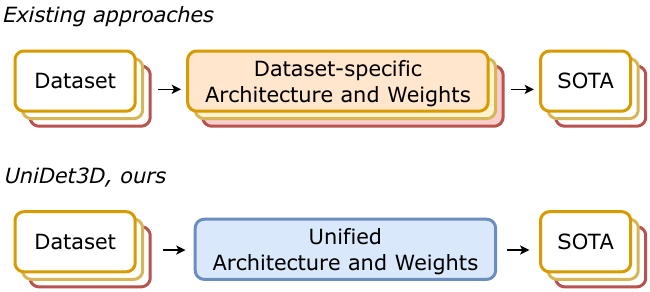}
    \caption{Existing 3D object detection methods use different architectures and weights to achieve state-of-the-art metrics on different datasets. We propose \ours{} trained single time on a mixture of datasets and achieving even better results.}
    \label{fig:teaser}
\end{figure}

State-of-the-art 3D object detection accuracy in indoor benchmarks is achieved via classic supervised training on categorical labels. Not only do they struggle to generalize to new visually distinct scenes and unseen objects -- but handling novel categories remains an unresolved issue.

In 2D object detection, training using data from other sources rather than the target domain is a fruitful direction being actively investigated. The most common and basic way is pretraining with diverse and voluminous out-of-domain data, followed by fine-tuning using in-domain data. Another paradigm implies training jointly on a mixture of in-domain and out-of-domain data. Similarly, we address the generalization of 3D object detection across domains represented in indoor 3D datasets. 

Creating a multi-dataset 3D object detection method can be decomposed into four sub-tasks.

First, the \textbf{network architecture} should be carefully designed so that handling data from different sources should not impose a major computational overhead. We claim a novel detection architecture as one of our major contributions. Currently, the best scores on each indoor benchmark are achieved by dataset-specific methods: sparse convolutional TR3D (on S3DIS)~\cite{rukhovich2023tr3d},  transformer-based V-DETR (on ScanNet)~\cite{shen2023v-detr}. On the contrary, we design a unified approach based on a pure self-attention encoder architecture without positional encoding and cross-attention.

Second, \textbf{training datasets} representing different domains should be properly chosen and mixed. We mix up long-lasting and well-known ScanNet~\cite{dai2017scannet}, S3DIS~\cite{armeni2016s3dis} and ARKitScenes~\cite{baruch2021arkitscenes}, and enrich them with smaller-scale MultiScan~\cite{mao2022multiscan}, 3RScan~\cite{wald20193rscan}, and ScanNet++~\cite{yeshwanthliu2023scannetpp}. 

Third, output data should be transformed into a label space shared across multiple datasets. To this end, we explore different ways of \textbf{merging category labels} across datasets with inconsistent annotations. 

Finally, the \textbf{multi-dataset training procedure} should be set up for robust performance in all domains -- rather than compromising the quality in most cases in favor of certain specific scenarios. To identify the best design choices, we experiment with several training strategies and show that joint training ensures higher scores on test splits of all datasets in the mixture.

\section{Related Work}

\subsection{3D Detection Architectures}
Existing 3D object detection methods can be categorized into voting-based, expansion-based, and transformer-based. The proposed \ours{} falls into the latter group, still we briefly overview both voting-based and expansion-based methods, since we include them in our quantitative comparison.

\paragraph{Voting-based} methods~\cite{qi2019votenet, chen2020hgnet, engelmann20203d-mpa, xie2020mlcvnet, cheng2021brnet, zhang2020h3dnet, xie2021venet, wang2022rbgnet, zheng2022hyperdet3d, zhu2024spgroup3d} pioneered the field, with VoteNet~\cite{qi2019votenet} being the first method that introduced point voting for 3D object detection. Subsequent methods mainly follow the line of extending VoteNet with additional modules and tricks to improve detection quality.
The latest work in this row, SPGroup3D~\cite{zhu2024spgroup3d}, exploits superpoint clustering, which has already proved itself to be beneficial for 3D instance segmentation~\cite{kolodiazhnyi2024oneformer3d}. Yet, using superpoints is not limited to voting-based approaches, and we also cluster an input point cloud into superpoints in our transformer-based \ours.

\paragraph{Expansion-based} methods~\cite{gwak2020gsdn, rukhovich2022fcaf3d, rukhovich2023tr3d, wang2022cagroup3d} generate virtual center features from surface features using a generative sparse decoder, and predict high-quality 3D region proposals. GSDN~\cite{gwak2020gsdn} adapts fully convolutional architecture for 3D object detection. FCAF3D~\cite{rukhovich2022fcaf3d} proposes anchor-free proposal generation, while TR3D~\cite{rukhovich2023tr3d} achieves real-time inference with a lightweight generative decoder. CAGroup3D~\cite{wang2022cagroup3d} improves the results of FCAF3D by running the second refinement stage.

\paragraph{Transformer-based} methods~\cite{misra20213detr, liu2021group-free, wang2024uni3detr, shen2023v-detr} dominate 3D object detection. Following the seminal Group-Free~\cite{liu2021group-free} work, they first extract point cloud features with a sparse-convolutional backbone and then predict objects from input queries with a transformer decoder through cross-attending to the backbone features. V-DETR~\cite{shen2023v-detr} upgrades Group-Free with a vertex relative positional encoding. Uni3DETR~\cite{wang2024uni3detr} extends over indoor and outdoor datasets. 3DETR~\cite{misra20213detr} replaces the backbone with a transformer encoder, making the entire network transformer-based. Overall, a decent part of progress in transformer-based 3D object detection is attributed to sophisticated architectures, elaborated positional encoding, and non-trivial interaction between modules. Besides, existing methods use computationally extensive Hungarian matching to assign predicted bounding boxes to ground truth ones during the training. 

On the contrary, we use a simple self-attention encoder architecture without positional encoding and cross-attention that are typically needed in the decoder part. We also replace Hungarian matching with a lightweight effective alternative. By designing \ours{} model, we follow a plug-and-play paradigm, so that each component can be easily replaced and tailored to user limitations and requirements. 

\subsection{Multi-dataset Object Detection}

Most existing object detection methods are trained on a single dataset, so that both the volume of data and semantics diversity are limited. Recently, training object detection on multiple datasets~\cite{cai2022bigdetection, shi2021multi, zhao2020object, zhou2022simple} has proved to boost the model quality, generalization ability, and robustness in the 2D domain. Different strategies of joining input sources and heterogeneous label spaces have been proposed so far, e.g., recent works~\cite{meng2023detectionhub, wang2023unidetector} leverage large language models to handle an open set of categories via representing them using natural language. Several attempts have been made to address multi-dataset training of 3D object detection in outdoor scenarios~\cite{zhang2023uni3d, soum2023mdt3d}, either in LIDAR point clouds or monocular images~\cite{brazil2023omni3d, li2024unimode}. We argue that outdoor-targeted approaches taking benefits from large-scale annotated datasets cannot be straightforwardly adapted to handle orders of magnitude smaller collections of indoor data. In this paper, we investigate multi-dataset training of 3D object detection in the indoor domain.

\section{Multi-dataset 3D Detection Training}

3D object detection aims to predict a location $b_i \in \mathbb{R}^7$ and a class-wise detection score $p_i \in \mathbb{R}^{|L|}$ for each object $i$ in a point cloud $P$. The detection score denotes confidence for a bounding box to belong to an object with label $c \in L$, where $L$ is the set of all classes (label space) of a dataset $\mathcal{D}$.

Many 3D object detection methods are trained and tested using the ScanNet dataset~\cite{dai2017scannet}, which contains balanced annotations for 18 common object classes; making training relatively simple. Training on ScanNet usually implies straightforwardly minimizing a loss $\ell$, e.g. a bounding box-level log-likelihood, over a sampled point cloud $\hat P$ and annotated 3D bounding boxes $\hat B$ from the dataset $\mathcal{D}$:
\begin{equation}
    \min_{\Theta} \mathbb{E}_{(\hat P, \hat B) \sim \mathcal{D}} \left[\ell(\mathcal{M}(\hat P; \Theta), \hat B)\right].
\end{equation}
Here, $\hat B$ contains class-specific box annotations. The loss $\ell$ is defined on two sets of bounding boxes, predicted and ground truth ones, being matched based on the overlap criterion.

\begin{figure}[h!]
    \centering
        \includegraphics[width=1.\linewidth]{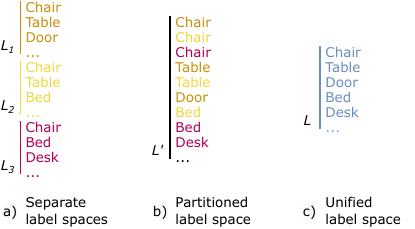}
    \caption{Three common ways of handling heterogeneous label spaces for training. The partitioned scheme implies using a separate classification head for each dataset. \ours{} follows the unified scheme, using the same de-duplicated set of labels during both the training and inference.}
    \label{fig:labels}
\end{figure}

Let us now consider training a 3D object detection model on $K$ datasets $\mathcal{D}_1, \mathcal{D}_2,\ldots$, having label spaces $L_1, L_2, \ldots$, respectively. The naive solution is to train a \textit{separate} model $M_k$ on a dataset $D_k$ using a dataset-specific loss $\ell_k$:
\begin{equation}
    \min_{\Theta} \mathbb{E}_{(\hat P, \hat B) \sim \mathcal{D}_k} \left[\ell_k(\mathcal{M}_k(\hat P; \Theta), \hat B)\right].
\end{equation}
However, training a single common model instead of several dataset-specific models can boost the performance for all datasets. In 2D object detection, this is achieved using a \textit{partitioned} label space~\cite{zhou2022simple}, which is equivalent to training $K$ models $\mathcal{M}_1, \ldots, \mathcal{M}_K$ with the same architecture $\mathcal{M}$ but the last dataset-specific classification layer. The common model is trained by minimizing the $K$ dataset-specific losses:
\begin{equation}
    \min_{\Theta} \mathbb{E}_{\mathcal{D}_k}\left[\mathbb{E}_{(\hat P, \hat B) \sim \mathcal{D}_k}\left[ \ell_k(\mathcal{M}_k(\hat P; \Theta), \hat B)\right]\right].
\end{equation}

Still, when using the partitioned label space, probabilities of classes from different datasets are estimated regardless of the similarity of these classes, e.g. probabilities for a chair category in the ScanNet and ARKitScenes datasets are predicted separately and independently (see Fig). While this may allow for better per-dataset scores, it complicates the interpretation of results for the end user, since per-dataset predictions should be somehow aggregated. This observation naturally leads to the \textit{unified} scheme, which is combining all labels of all datasets into a dataset $\mathcal{D} = \mathcal{D}_1 \cup  \mathcal{D}_2 \cup \ldots$, and uniting the label spaces $L=L_1 \cup L_2 \cup \ldots$. Similar labels get merged, making the common label space unambiguous. The optimization procedure remains the same as for default training on a single dataset:
\begin{equation}
    \min_{\Theta} \mathbb{E}_{(\hat P, \hat B) \sim \mathcal{D}_1 \cup  \mathcal{D}_2 \cup \ldots} \left[\ell(\mathcal{M}(\hat P; \Theta), \hat B)\right].
\end{equation}
In an empirical study below, we show that the unified scheme supersedes the partitioned one not only regarding simplicity, interpretability, and fewer training parameters but also delivers higher accuracy.

\section{3D Detection Network}

The overall scheme of \ours{} is depicted in Fig. Given a point cloud, a sparse 3D U-Net network extracts point-wise features. In parallel, superpoints are obtained through unsupervised clustering. Then, point features are aggregated within each superpoint by simple averaging (or superpoint pooling), giving superpoint features. Superpoint features are passed as queries to a vanilla transformer encoder. The encoder outputs are processed with two separate MLPs, one estimating regression parameters of objects' bounding boxes, and another predicting class probabilities in multi-dataset shared label space.

\begin{figure*}[h!]
    \centering
        \includegraphics[width=1.\linewidth]{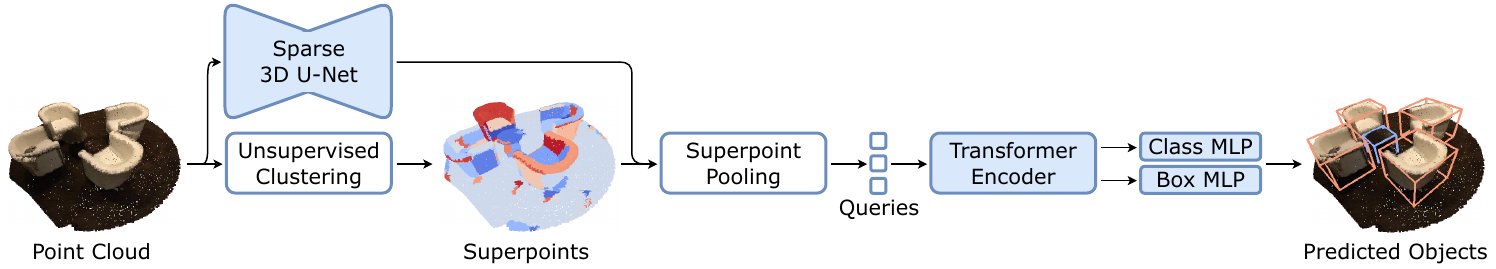}
    \caption{Overview of the proposed method. \ours{} takes the point cloud as an input, and extracts point features using a sparse 3D U-Net network. Point features are averaged across superpoints in the superpoint pooling. Aggregated features serve as input queries to a vanilla transformer encoder. Finally, 3D bounding boxes are derived from encoder outputs with a box MLP and class MLP, where box MLP estimates the location of a 3D bounding box w.r.t. the mass center of the superpoint, and class MLP outputs probabilities of object classes in the unified label space.}
    \label{fig:scheme}
\end{figure*}

\begin{figure}
    \centering
        \includegraphics[width=1.\linewidth]{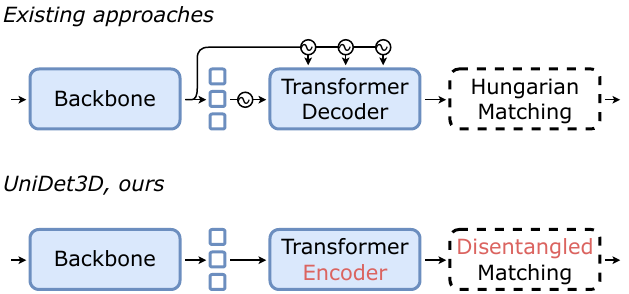}
    \caption{Comparison with existing transformed-based 3D object detection methods. We introduce encoder-only transformer architecture w/o positional encoding for queries or attention layers. This allows us to change unstable Hungarian matching for a simpler disentangled scheme.}
    \label{fig:transformer}
\end{figure}

\subsection{Backbone and Pooling}

\paragraph{3D U-Net.} Assuming that an input point cloud contains $N$ points, the input can be formulated as $P \in \mathbb{R}^{N \times 6}$. Each 3D point is parameterized with three colors \textit{r}, \textit{g}, \textit{b}, and three coordinates \textit{x}, \textit{y}, \textit{z}. Following \cite{choy2019minkowski}, we voxelize the point cloud and use a U-Net-like backbone composed of sparse 3D convolutions to extract point-wise features $P' \in \mathbb{R}^{N \times C}$.

\paragraph{Superpoint pooling.} To build an end-to-end framework, we directly feed point-wise features $P' \in \mathbb{R}^{N \times C}$ into superpoint pooling layer based on pre-computed superpoints~\cite{landrieu2018superpoints}. The superpoint pooling layer obtains superpoint features $S \in \mathbb{R}^{M \times C}$ via average pooling over those point-wise ones inside each superpoint. Without loss of generality, we suppose that there are $M$ superpoints computed from the input point cloud. Notably, the superpoint pooling layer downsamples the input point cloud to hundreds of superpoints, which significantly reduces the computational overhead of subsequent processing and optimizes the representation capability of the entire network.

\subsection{Transformer Encoder}

Backbone features after superpoint pooling are processed with a transformer encoder network. 
This network takes $M$ queries as an input and outputs $M$ object proposal features. 

Existing transformer-based 3D object detection methods~\cite{misra20213detr, liu2021group-free, wang2024uni3detr, shen2023v-detr} exploit different techniques to establish connections between queries and points in an input point cloud. In particular, queries are initialized via the farthest point sampling. Then, positional encoding is added to preserve spatial information, and queries are cross-attended to points with positional guidance in a \textit{decoder} part of the network. On the contrary, we use a vanilla transformer \textit{encoder} without bells and whistles (Fig.~\ref{fig:transformer}). 

Overall, we employ a simple, elegant transformer architecture based on self-attention between input queries solely. Our experiments show that positional encoding is redundant, bringing negligible to no accuracy improvement at the cost of extra computations and more complex design. 

\subsection{Head}

The head takes $M$ object proposals as inputs and produces a single 3D bounding box and a class label for each proposal, estimated via linear layers.
The classification layer (\textit{class MLP} in the Fig.~\ref{fig:scheme}) outputs probabilities of $|L|$ object classes. The parameters of 3D bounding boxes are regressed with a \textit{box MLP}. Each 3D bounding box is represented in the form of an 8-value vector~\cite{rukhovich2022fcaf3d}, where the first six values denote positive distances from the proposal coordinate to the faces of a predicted 3D bounding box, and the last two values define a rotation angle. The distances to the faces are given w.r.t. the mass center of the superpoint, making it the only positional information used to predict 3D bounding boxes.

\subsection{Training}

To train a transformer-based method end-to-end, we need to define a cost function between queries and ground truth objects, develop a matching strategy that minimizes this cost function, and formulate a loss function being applied to the matched pairs.

\paragraph{Cost function.} We use a pairwise matching cost $\mathcal{C}_{ik}$ to measure the similarity of the $i$-th proposal and the $k$-th ground truth. $\mathcal{C}_{ik}$ is derived from a classification probability and predicted bounding box:
\begin{equation}
    \label{equ:C}
    \mathcal{C}_{ik} = -\lambda \cdot  p_{i,c_k}+DIoU(b_i, \hat{b}_k),
\end{equation}
where $p_{i,c_k}$ indicates the probability of $i$-th proposal belonging to the $c_k$ semantic category. $b_i$ and $\hat{b}_k$ stands for predicted and ground truth bounding boxes. The distance between ground truth and predicted boxes is measured with the DIoU function as in TR3D~\cite{rukhovich2023tr3d}. In our experiments, we use $\lambda = 0.25$.

\paragraph{Matching.} Previous transformer-based 3D object detection methods~\cite{misra20213detr, liu2021group-free, wang2024uni3detr, shen2023v-detr} use bipartite matching based on a Hungarian algorithm \cite{kuhn1955hungarian}. This common approach has a major drawback: an excessive number of meaningful matches between proposals and ground truth objects makes the training process long-lasting and unstable. To overcome this issue, we adapt the disentangled matching scheme introduced in recent 3D instance segmentation work~\cite{kolodiazhnyi2024oneformer3d}.

Since an object query is initialized with superpoint features, this object query can be unambiguously matched with this superpoint. We assume that a superpoint can belong only to one object, which gives a correspondence between a superpoint, an object query, an object proposal derived from this object query, and a ground truth object. 

We assign each object with three nearest superpoints, so, to get a bipartite matching, we only need to filter out extra superpoints matched to the same object. This task reformulation simplifies cost function optimization, as we can set the most weights in a cost matrix to infinity: 
\begin{equation}
\mathcal{\hat{C}}_{ik}=\left\{  
    \begin{array}{rl} 
    \mathcal{C}_{ik} & \text{if $i$-th superpoint $\in$ $k$-th object} \\
    +\infty & \text{otherwise}
    \end{array}
\right.
\end{equation}

\paragraph{Loss.} After matching proposals with ground truth instances, instance losses can finally be calculated. Classification errors are penalized with a cross-entropy loss $\mathcal{L}_{cls}$. Besides, for each match between a proposal and a ground truth object, we compute the regression loss $\mathcal{L}_{reg}$ as a DIoU function between predicted and ground truth boxes.

The total loss $\mathcal{L}$ is formulated as:
\begin{equation}
\label{equ:L}
\mathcal{L}=\beta \cdot \mathcal{L}_{cls}+\mathcal{L}_{reg}
\end{equation}
where $\beta = 0.5$.

\section{Experiments}

\subsection{Datasets}

\begin{table}[h!]
    \centering
    \begin{tabular}{lccc}
        \toprule
        \multirow{2}{*}{Dataset} & \# Train. & \# Val. & \multirow{2}{*}{\# Classes} \\
        & Scenes & Scenes & \\
        \midrule
        ScanNet & 1201 & 312 & 18 \\
        ARKitScenes & 4493 & 549 & 17\\
        S3DIS & 204 & 68 & 5 \\
        MultiScan & 174  & 42 & 17 \\
        3RScan & 385 & 47 & 18 \\
        ScanNet++ & 230 & 50 & 84 \\
        \midrule
        Overall & 6687 & 1068 & 99 \\ 
        \bottomrule
    \end{tabular}
    \caption{Quantitative statistics of indoor datasets in our mixture. ScanNet and ARKitScenes are relatively large-scale, while S3DIS, MultiScan, 3RScan, and ScanNet++ are times smaller.}
    \label{tab:datasets}
\end{table}

We evaluate our method on six real-world indoor benchmarks: ScanNet~\cite{dai2017scannet}, ARKitScenes~\cite{baruch2021arkitscenes}, S3DIS~\cite{armeni2016s3dis}, MultiScan~\cite{mao2022multiscan}, 3RScan~\cite{wald20193rscan}, ScanNet++~\cite{yeshwanthliu2023scannetpp}. For methodological purity, we do not add single-view RGB-D datasets such as SUN RGB-D~\cite{song2015sunrgbd} to the mixture but only use datasets containing multi-view 3D reconstructions. In the absence of ground truth 3D bounding boxes, we calculate axis-aligned bounding boxes from 3D instance labels through a standard approach \cite{qi2019votenet}.

\paragraph{ScanNet}~\cite{dai2017scannet} contains 1513 reconstructed 3D indoor scans with per-point instance and semantic labels of 18 categories. The training subset consists of 1201 scans, while 312 scans are used for validation.

\begin{table*}[h!]
\setlength{\tabcolsep}{3pt}
\centering
\resizebox{\linewidth}{!}{\begin{tabular}{lllcccccccccccc}
\toprule
& \multirow{2}{*}{Method} & \multirow{2}{*}{Venue} & \multicolumn{2}{c}{ScanNet} & \multicolumn{2}{c}{ARKitScenes} & \multicolumn{2}{c}{S3DIS} & \multicolumn{2}{c}{MultiScan} & \multicolumn{2}{c}{3RScan} & \multicolumn{2}{c}{ScanNet++} \\
& & & mAP\textsubscript{25} & mAP\textsubscript{50} & mAP\textsubscript{25} & mAP\textsubscript{50} & mAP\textsubscript{25} & mAP\textsubscript{50} & mAP\textsubscript{25} & mAP\textsubscript{50} & mAP\textsubscript{25} & mAP\textsubscript{50} & mAP\textsubscript{25} & mAP\textsubscript{50} \\
\midrule
\multicolumn{14}{l}{\textit{Best result}}  \\
& VoteNet & ICCV’19 & 58.6 & 33.5 & 35.8 \\
& HGNet & CVPR’20 & 61.3 & 34.4 \\
& GSDN & ECCV’20 & 62.8 & 34.8 & & & 47.8 & 25.1 \\
& 3D-MPA & CVPR’20 & 64.2 & 49.2 \\
& MLCVNet & CVPR’20 & 64.5 & 41.4 & 41.9 \\
& 3DETR & ICCV’21 & 65.0 & 47.0 \\
& BRNet & CVPR’21 & 66.1 & 50.9 \\
& H3DNet & ECCV’20 & 67.2 & 48.1 & 38.3 \\
& VENet & ICCV’21 & 67.7 \\
& Group-Free & ICCV’21 & 69.1 & 52.8 \\
& RBGNet & CVPR’22 & 70.6 & 55.2 \\
& HyperDet3D & CVPR’22 & 70.9 & 57.2 \\
& FCAF3D & ECCV’22  & 71.5 & 57.3 & & & 66.7 & 45.9 & \reimpl{53.8} & \reimpl{40.7} & \reimpl{60.1} & \reimpl{42.6} & \reimpl{22.3} & \reimpl{11.4} \\
& Uni3DETR & NIPS’23 & 71.7 & 58.3 & & & 70.1 & 48.0 \\
& TR3D & ICIP’23 & 72.9 & 59.3 & & & 74.5 & 51.7 & \reimpl{56.7} & \reimpl{42.3} & \reimpl{62.3} &  \reimpl{45.4} & \reimpl{26.2} & \reimpl{14.5} \\
& SPGroup3D & AAAI’24 & 74.3 & 59.6 & & & 69.2 & 47.2 \\
& CAGroup3D & NIPS’22 & 75.1 & 61.3 \\
& V-DETR & ICLR’24 & 77.4 & 65.0 \\
& \textbf{UniDet3D} & & \textbf{77.5} & \textbf{66.1} & \textbf{61.3} & \textbf{47.1} & \textbf{75.2} & \textbf{60.8} & \textbf{64.2} & \textbf{51.6} & \textbf{64.7} & \textbf{48.6} & \textbf{26.4} & \textbf{17.2} \\
\bottomrule
\multicolumn{14}{l}{\textit{Average across 25 trials}} \\
& Group-Free & ICCV’21 & 68.6 & 51.8 \\
& RBGNet & CVPR’22 & 69.9 & 54.7 \\
& FCAF3D & ECCV’22 & 70.7 & 56.0 & & & 64.9 & 43.8 & \reimpl{52.5} & \reimpl{39.2} & \reimpl{59.6} & \reimpl{40.4} & \reimpl{21.4} & \reimpl{11.0} \\
& TR3D & ICIP’23 & 72.0 & 57.4 & & & 72.1 & 47.6 & \reimpl{55.0} & \reimpl{41.2} & \reimpl{61.5} & \reimpl{44.2} & \reimpl{24.3} & \reimpl{13.9} \\
& SPGroup3D & AAAI’24 & 73.5 & 58.3 & & & 67.7 & 43.6 \\
& CAGroup3D & NIPS’22 & 74.5 & 60.3 \\
& V-DETR & ICLR’24 & 76.8 & 64.5 \\
& \textbf{UniDet3D} & & \textbf{77.1} & \textbf{65.2} & \textbf{60.2} & \textbf{46.0}  & \textbf{73.3} & \textbf{57.9} & \textbf{62.4} & \textbf{50.8} & \textbf{62.1} & \textbf{45.6} & \textbf{24.4} & \textbf{16.3} \\
\bottomrule
\end{tabular}}
\caption{Comparison of the detection methods on 6 datasets: ScanNet, S3DIS, ARKitScenes, MultiScan, 3RScan, and ScanNet++. Our \ours{} trained jointly on 6 datasets sets the new state-of-the-art in all benchmarks. Results obtained by running existing methods on the novel datasets are marked \reimpl{gray}.}
\label{tab:results}
\end{table*}

\paragraph{ARKitScenes}~\cite{baruch2021arkitscenes} consists of 5042 scans of 1661 venues captured using a tablet with an online ARKit tracking system. This is the only dataset in the list labeled with oriented bounding boxes. We use the official training and validation splits of 4493 and 549 scans, respectively. 

\paragraph{Stanford Large-Scale 3D Indoor Spaces (S3DIS)}~\cite{armeni2016s3dis} contains scans of 272 rooms, annotated with instance and semantic labels of five furniture categories. We use the official \textit{Area 5} split, where 68 rooms serve for validation, and 204 rooms comprise the training subset.

\paragraph{MultiScan}~\cite{mao2022multiscan} is a small yet extensively labeled RGB-D dataset of 273 scans of 117 indoor scenes with 11K objects, primarily intended for part mobility estimation. It contains per-frame camera poses, textured 3D surface meshes, and fine object-level semantic labels.

\paragraph{3RScan}~\cite{wald20193rscan} is designed as a benchmark for temporal visual analysis, e.g., change detection or visual localization. It features 1482 3D reconstructions of 478 scenes alongside calibrated RGB-D sequences, textured 3D meshes and instance and semantic annotations.  

\paragraph{ScanNet++}~\cite{yeshwanthliu2023scannetpp} is an instance segmentation and novel-view synthesis benchmark. It contains 450 RGB-D sequences recorded with an iPhone and 3D scans captured using a laser scanner with sub-millimeter resolution and annotated with long-tail semantics.

\subsection{Evaluation}

For all datasets, we use mean average precision (mAP) under IoU thresholds of 0.25 and 0.5 as a metric.

We upper-limit the number of points in an input point cloud by $N=100000$ points, as proposed in ~\cite{rukhovich2022fcaf3d, rukhovich2023tr3d}. Since these points are sampled randomly, both training and evaluation procedures are randomized. To obtain statistically significant results, we run training 5 times and test each trained model 5 times independently. We report the best and average metrics across 5 $\times$ 5 trials: this allows comparing \ours{} to the 3D object detection methods that report either a single best or an average value.

\begin{table*}[h!]
\setlength{\tabcolsep}{3pt}
\centering
\begin{tabular}{llcccccccccccc}
\toprule
& \multirow{2}{*}{Label Space} & \multicolumn{2}{c}{ScanNet} & \multicolumn{2}{c}{ARKitScenes} & \multicolumn{2}{c}{S3DIS} & \multicolumn{2}{c}{MultiScan} & \multicolumn{2}{c}{3RScan} & \multicolumn{2}{c}{ScanNet++} \\
& & mAP\textsubscript{25} & mAP\textsubscript{50} & mAP\textsubscript{25} & mAP\textsubscript{50} & mAP\textsubscript{25} & mAP\textsubscript{50} & mAP\textsubscript{25} & mAP\textsubscript{50} & mAP\textsubscript{25} & mAP\textsubscript{50} & mAP\textsubscript{25} & mAP\textsubscript{50} \\
\midrule
\multicolumn{3}{l}{\textit{from scratch}} \\
& separate & 77.0 & 65.0 & 59.6 & 45.7 & 57.2 & 39.7 & 46.1 & 33.1 & 45.1 & 31.4 & 21.6 & 12.2 \\
\midrule
\multicolumn{3}{l}{\textit{joint training}} \\
& partitioned  & 77.0 & 65.1 & 59.8 & 45.8 & 71.2 & 56.2 & 62.0 & 50.5 & \textbf{62.6} &  45.4 & 24.1 & 16.0 \\
& unified & \textbf{77.1} & \textbf{65.2} & \textbf{60.2} & \textbf{46.0}  & \textbf{73.3} & \textbf{57.9} & \textbf{62.4} & \textbf{50.8} & 62.1 & \textbf{45.6} & \textbf{24.4} & \textbf{16.3} \\
\bottomrule
\end{tabular} 
\caption{Scores (average across 25 trials) obtained using different label spaces. Expectedly, joint training is especially beneficial for small datasets. Switching from the partitioned (159 classes) to unified (99 classes) label space not only increases interpretability for an end user but also has a positive effect on overall accuracy, which is a valuable practical outcome.}
\label{tab:label-space}
\end{table*}

\begin{table}[h!]
\centering
\setlength{\tabcolsep}{2pt}
\resizebox{\linewidth}{!}{
\begin{tabular}{lccccccccc}
\toprule
& Scan- & ARKit-  & \multicolumn{2}{c}{S3DIS} & \multicolumn{2}{c}{MultiScan}  & \multicolumn{2}{c}{3RScan} \\
& Net & Scenes & mAP\textsubscript{25} & mAP\textsubscript{50} & mAP\textsubscript{25} & mAP\textsubscript{50}  & mAP\textsubscript{25} & mAP\textsubscript{50} \\
\midrule
\multicolumn{4}{l}{\textit{from scratch}} & & & \\
& & & 57.2 & 39.7 & 46.1 & 33.1 & 45.1 & 31.4 \\
\midrule
\multicolumn{3}{l}{\textit{fine-tuning}} & & & \\
& \cmark & \cmark  & 71.3 & 54.3 & 60.2 & 49.1 & 60.8 & 45.6 \\
\midrule
\multicolumn{3}{l}{\textit{joint training}} & & & \\
& \cmark &  & 72.0 & 55.3 & 59.0 & 46.2 & 59.8 & 42.0 \\
& & \cmark &  65.5 & 48.3 & 46.5 & 34.7 & 55.4 & 40.5  \\
& \cmark & \cmark  & \textbf{73.3} & \textbf{57.9} & \textbf{62.4} & \textbf{50.8} & \textbf{62.1} & \textbf{45.6} \\
\bottomrule
\end{tabular}}
\caption{Scores (average across 25 trials) obtained on the S3DIS, MultiScan, and 3RScan test splits after either training from scratch on the train splits on the respective datasets, using pre-training, or joint training. The joint training on the mixture of larger ScanNet and ARKitScenes datasets is the most beneficial.}
\label{tab:pretraining}
\end{table}

\subsection{Implementation Details}

We implement \ours{} in the mmdetection3d~\cite{2020mmdetection3d} framework. All training details are the same as in OneFormer3D~\cite{kolodiazhnyi2024oneformer3d}, particularly, we use AdamW optimizer with an initial learning rate of 0.0001, weight decay of 0.05, batch size of 8, and polynomial scheduler with a base of 0.9 for 1024 epochs. We apply the standard augmentations: horizontal flipping, random rotations around the z-axis, and random scaling. During the training, we assign a ground truth object to the three nearest superpoints. Since during the inference we seek for one-to-one matching, we suppress redundant superpoints using NMS. No test-time augmentations are applied during the inference time. All experiments are conducted using a single NVidia V100.

\subsection{Comparison to Prior Work}

We compare \ours{} against various 3D object detection methods. According to the Tab.~\ref{tab:results}, \ours{} consistently outperforms the competitors not only in the \textit{best} but also in the \textit{average} scores, which indicates the statistical significance of results. In the well-known ScanNet and S3DIS benchmarks, \ours{} sets state-of-art results, superseding second-best scores by at least +1 mAP\textsubscript{50} on ScanNet and impressive +9.1 mAP\textsubscript{50} on S3DIS. To obtain reference values for smaller datasets MultiScan, 3RScan, and ScanNet++, we train and evaluate FCAF3D~\cite{rukhovich2022fcaf3d} and TR3D~\cite{rukhovich2023tr3d}, two strong baselines with publicly available code. The observed improvement over these methods is especially tangible for MultiScan, where the gain is +7.5 mAP\textsubscript{25} and +9.3 mAP\textsubscript{50}.

\begin{table}[h!]
\centering
\resizebox{\linewidth}{!}{
\begin{tabular}{lccccc}
    \toprule
    \multirow{2}{*}{Method} & \multirow{2}{*}{PE} & \multirow{2}{*}{HM} & \multirow{2}{*}{mAP\textsubscript{25}} & \multirow{2}{*}{mAP\textsubscript{50}} & Inference \\
    & & & & & time, ms \\
    \midrule
    3DETR & \cmark & \cmark & 65.0 & 47.0 & 170 \\
    Group-Free & \cmark & \cmark & 69.1 & 52.8 & 157 \\
    Uni3DETR & \cmark & \cmark & 71.7 & 58.3 & 283 \\
    V-DETR & \cmark & \cmark & 77.4 & 65.0 & 240 \\
    \midrule
    \multirow{3}{*}{\textbf{UniDet3D}} & \cmark & & 77.4 & 66.0 & 233 \\
    & & \cmark & 75.2 & 64.5 & 224 \\
    & & & \textbf{77.5} & \textbf{66.1} & 224 \\
    \bottomrule
\end{tabular}}
\caption{Comparison of transformed-based methods on the ScanNet validation split, all trained on the ScanNet training split solely. PE is positional encoding, HM is Hungarian matching (applied only during the training). \ours{} without PE and HM hits the highest scores.}
\label{tab:enconding}
\end{table}

\subsection{Ablation Studies}

\paragraph{Training schemes.}

To emulate real usage, we consider small S3DIS, MultiScan, and 3RScan as target datasets, and leverage large ScanNet and ARKitScenes as sources of extra training data. In Tab.~\ref{tab:pretraining}, we compare three training schemes: 
\begin{itemize}
\item training \textit{from scratch} on target dataset;
\item \textit{fine-tuning} after pre-training on a mixture of ScanNet and ARKitScenes;
\item \textit{joint training} on a mixture of ScanNet and/or ARKitScenes and the target dataset.
\end{itemize}

While transformer-based approaches dominate on large-scale datasets, they cannot train sufficiently on limited data -- which is the case when using extra data and more elaborate training schemes appears to be the most profitable. Respectively, \ours{} easily outperforms both transformer and non-transformer methods on the large ScanNet, but is notably inferior to convolutional baselines TR3D and FCAF3D, when trained \textit{from scratch} on S3DIS, MultiScan, or 3RScan.

After simple \textit{fine-tuning}, our model surpasses baseline approaches, which evidences our unified architecture to effectively adapt to target domains after learning general concepts from the voluminous mixture of training datasets. This result is valuable for practitioners seeking a customizable approach that can be trained quickly under limited computational powers. \textit{Joint training} adds extra +3.6 and +1.7 mAP\textsubscript{50} on S3DIS and MultiScan, respectively. Expectedly, the amount and variety of training data also matter: using both ScanNet and ARKitScenes ensures higher accuracy than using them solely.

\paragraph{Merging different label spaces.}

The benefits of joint training are fully revealed for small datasets, and so is the difference between partitioned and unified label spaces. According to Tab~\ref{tab:label-space}, unifying label space improves the overall quality over the partitioned label space and brings +1.7 mAP\textsubscript{50} on S3DIS. Taking the better interpretability of unified classes and the smaller size of the classification layer (99 unified classes against 159 in the partitioned label space), we can claim the unified label space as a preferred option. Not only is this an interesting experimental finding, but a useful feature for real-world applications.

\paragraph{Positional encoding and Hungarian matching.}

In this study, we measure the effect of positional encoding and matching strategy on the model's performance. 

To match randomly initialized queries and point cloud features, transformer-based methods use positional encoding and cross-attention. Since our superpoint-induced query initialization strategy preserves spatial information, the need for adding positional encoding is questionable. Apart from that, \ours{}'s query initialization allows employing \textit{disentangled matching} instead of costly \textit{Hungarian matching}.

To ensure competitive comparison, we implement vertex relative positional encoding proposed in the previous state-of-the-art V-DETR~\cite{shen2023v-detr}. As seen in Tab.~\ref{tab:enconding}, \ours{} trained without positional encoding and Hungarian matching achieves the highest detection accuracy on ScanNet among transformer-based methods. In the meantime, eliminating positional encoding reduces time- and memory- footprint, so overall we can claim both transformer-specific parts to be redundant.

\section{Conclusion}

In this work, we proposed \ours{}, a 3D object detection model trained on a mixture of indoor datasets.
By unifying label spaces across datasets in the supervised joint training scheme, \ours{} generalizes to various indoor environments. The network architecture of the proposed method is built upon a vanilla transformer encoder, making the entire pipeline easy to use and adapt to user requirements. Extensive experiments prove \ours{} to deliver state-of-the-art results in 6 indoor benchmarks: ScanNet, ARKitScenes, S3DIS, MultiScan, 3RScan, and ScanNet++.

\appendix

\section{Ablation Studies}

\subsection{Number of Transformer Blocks}

Our experiments with varying number of transformer layers reveal that in general, the more blocks, the better. Yet, after reaching a threshold value of 6 layers, the computational footprint keeps increasing while the accuracy remains on a plateau (see Tab.~\ref{tab:layers}). Accordingly, we use 6 transformer layers in all our experiments unless otherwise specified.

\begin{table}[b!]
\centering
\begin{tabular}{ccc}
    \toprule
    \# Layers & mAP\textsubscript{25} & mAP\textsubscript{50} \\
    \midrule
    0 & 68.4 & 52.2 \\
    1 & 72.4 & 55.5 \\
    2 & 74.1 & 59.1 \\
    3 & 74.3 & 63.2 \\
    4 & 76.1 & 64.3 \\
    5 & 76.7 & 64.6 \\
    \textbf{6} & \textbf{77.1} & \textbf{65.2} \\
    9 & 77.1 & 65.1 \\
    \bottomrule
\end{tabular}
\caption{Results of models with different numbers of transformer layers in \ours{}'s encoder on ScanNet. Scores are averaged across 25 trials. The model with 6 transformer layers is the most accurate.}
\label{tab:layers}
\end{table}

\begin{table}[hb!]
\centering
\begin{tabular}{lccc}
    \toprule
    \multirow{2}{*}{Backbone} & \multirow{2}{*}{mAP\textsubscript{25}} & \multirow{2}{*}{mAP\textsubscript{50}} & Inference \\
    & & & time, ms \\
    \midrule
    Minkowski & 74.1 & 61.5 & 238 \\
    SpConv & \textbf{77.1} & \textbf{65.2} & \textbf{224} \\
    \bottomrule
\end{tabular}
\caption{Results of models with sparse backbones implemented using different 3D sparse convolutional backends on ScanNet. Scores are averaged across 25 trials. SpConv outperforms the Minkowski engine in accuracy and efficiency.}
\label{tab:backbone}
\end{table}

\begin{table}[ht!]
\centering
\begin{tabular}{llcc}
    \toprule
    Method & Venue & mAP\textsubscript{25} & mAP\textsubscript{50} \\
    \midrule
    GSPN & CVPR’19 & 30.6 & 17.7 \\
    DyCo3D & CVPR’21 & 58.9 & 45.3 \\
    PointGroup & CVPR’20 & 61.5 & 48.9 \\
    SSTNet & ICCV’21 & 62.5 & 52.7 \\
    3D-MPA & CVPR’20 & 64.2 & 49.2 \\
    HAIS & ICCV’21 & 64.3 & 53.1 \\
    DKNet & ECCV’22 & 67.4 & 59.0 \\
    Mask3D & ICRA’23 & 71.0 & 56.6 \\
    SoftGroup & CVPR’22 & 71.6 & 59.4 \\
    QueryFormer & ICCV’23 & 73.4 & 61.7 \\
    OneFormer3D & CVPR’24 & 76.9 & 65.3\\
    \textbf{UniDet3D} & & \textbf{77.5} & \textbf{66.1} \\
    \bottomrule
\end{tabular}
\caption{3D object detection scores of \ours{} and 3D instance segmentation methods on ScanNet. Even trained on weaker box-level annotations, \ours{} outperforms 3D instance segmentation methods in detection accuracy.}
\label{tab:instance}
\end{table}

\subsection{Sparse Convolutional Backend}

From the practical point of view, the choice of backend is crucial, especially when it comes to efficient processing of complex unstructured inputs, such as 3D sparse data. Following OneFormer3D~\cite{kolodiazhnyi2024oneformer3d}, we compare backbones implemented using two popular 3D sparse convolutional backends: MinkowskiEngine~\cite{choy2019minkowski} and SpConv~\cite{spconv2022}. In our experiments, SpConv proved itself to be the best choice regarding both inference speed and 3D object detection quality (Tab. ~\ref{tab:backbone}).


\subsection{Comparison with 3D Instance Segmentation Methods}

We compare \ours{} with recent 3D instance segmentation methods ~\cite{yi2019gspn, he2021dyco3d, jiang2020pointgroup, liang2021sstnet, engelmann20203d-mpa, chen2021hais, wu2022dknet, schult2023mask3d, lu2023queryformer, kolodiazhnyi2024oneformer3d} in Tab. ~\ref{tab:instance}. During training, these methods benefit from per-point instance mask annotations, while \ours{} has access only to box-level annotations. Still, \ours{} delivers higher 3D object detection quality.

\section{Qualitative Results}

To give an intuition on how the detection scores relate to actual detection quality, we provide additional visualizations of ground truth and predicted boxes for point clouds from all six datasets: ScanNet, ARKitScenes, S3DIS, MultiScan, 3RScan, and ScanNet++ in Fig.~\ref{fig:visualization}. All visualizations are produced with a model trained once jointly on the six datasets.

\section{Class Names}

\ours{} is trained to predict labels in the unified label space of size 99, obtained by merging 6 per-dataset label spaces. Below, we list all labels in each dataset.

\paragraph{ScanNet} 18 classes: \textit{bathtub, bed, bookshelf, cabinet, chair, counter, curtain, desk, door, garbagebin, picture, refrigerator, showercurtrain, sink, sofa, table, toilet, window.}

\paragraph{ARKitScenes} 17 classes: \textit{bathtub, bed, cabinet, chair, dishwasher, fireplace, oven, refrigerator, shelf, sink, sofa, stool, stove, table, toilet, tv monitor, washer.}

\paragraph{S3DIS} 5 classes: \textit{board, bookcase, chair, sofa, table.}

\paragraph{MultiScan} 17 classes: \textit{backpack, bed, cabinet, chair, curtain, door, microwave, pillow, refrigerator, sink, sofa, suitcase, table, toilet, trash can, tv monitor, window.}

\paragraph{3RScan} 18 classes: \textit{bathtub, bed, bookshelf, cabinet, chair, counter, curtain, desk, door, garbagebin, picture, refrigerator, showercurtrain, sink, sofa, table, toilet, window. }

\paragraph{ScanNet++} 84 classes: \textit{backpack, bag, basket, bed, binder, blanket, blinds, book, bookshelf, bottle, bowl, box, bucket, cabinet, ceiling lamp, chair, clock, coat hanger, computer tower, container, crate, cup, curtain, cushion, cutting board, door, exhaust fan, file folder, headphones, heater, jacket, jar, kettle, keyboard, kitchen cabinet, laptop, light switch, marker, microwave, monitor, mouse, office chair, painting, pan, paper bag, paper towel, picture, pillow, plant, plant pot, poster, pot, power strip, printer, rack, refrigerator, shelf, shoe rack, shoes, sink, slippers, smoke detector, soap dispenser, socket, sofa, speaker, spray bottle, stapler, storage cabinet, suitcase, table, table lamp, tap, telephone, tissue box, toilet, toilet brush, toilet paper, towel, trash can, tv, whiteboard, whiteboard eraser, window.}

\begin{table*}
\setlength{\tabcolsep}{1pt}
\centering
\begin{tabular}{cccc}
   Ground Truth & Predicted & Ground Truth & Predicted \\
   \makecell[l]{\includegraphics[width=0.24\linewidth]{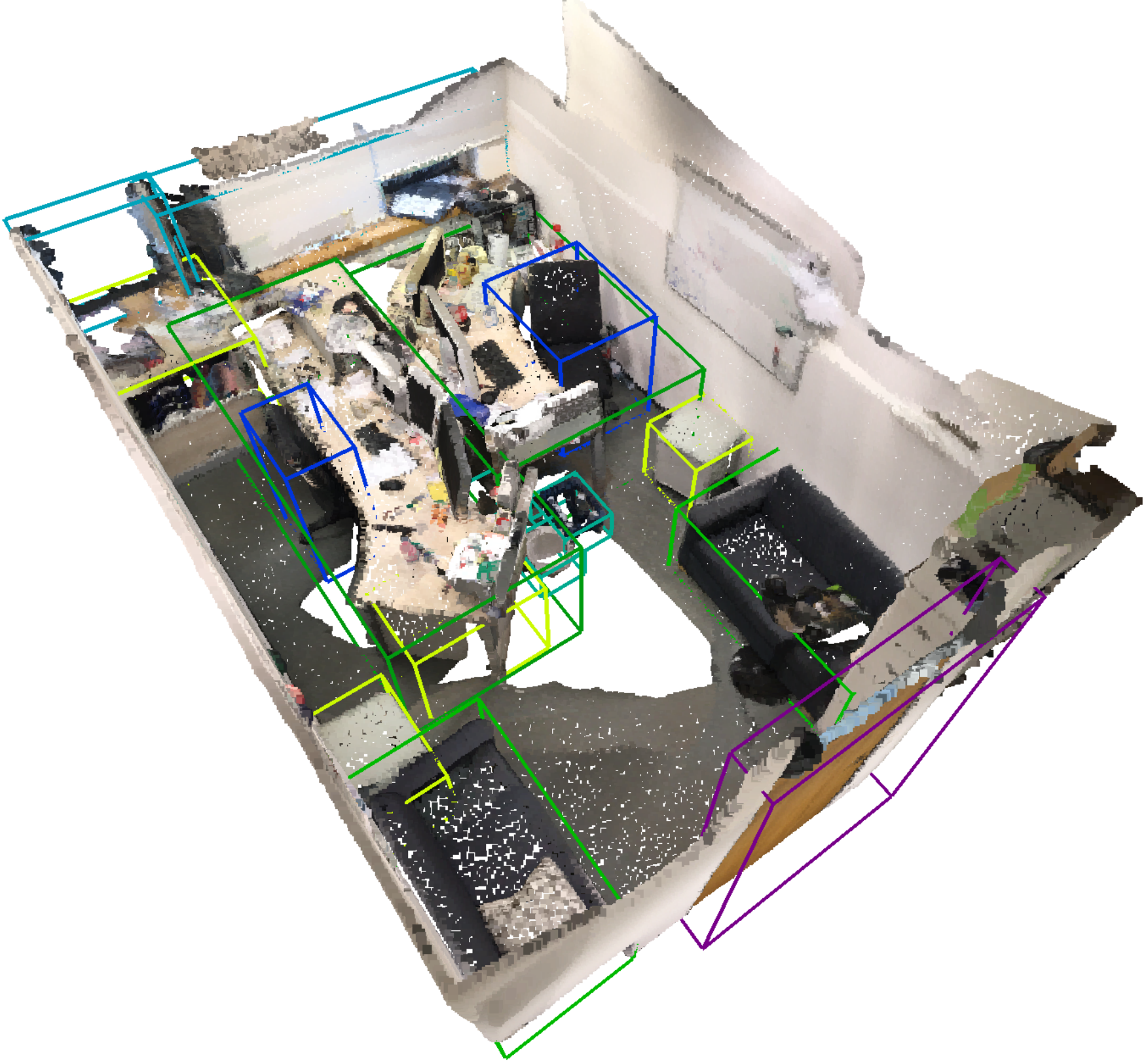}} &
   \makecell[l]{\includegraphics[width=0.24\linewidth]{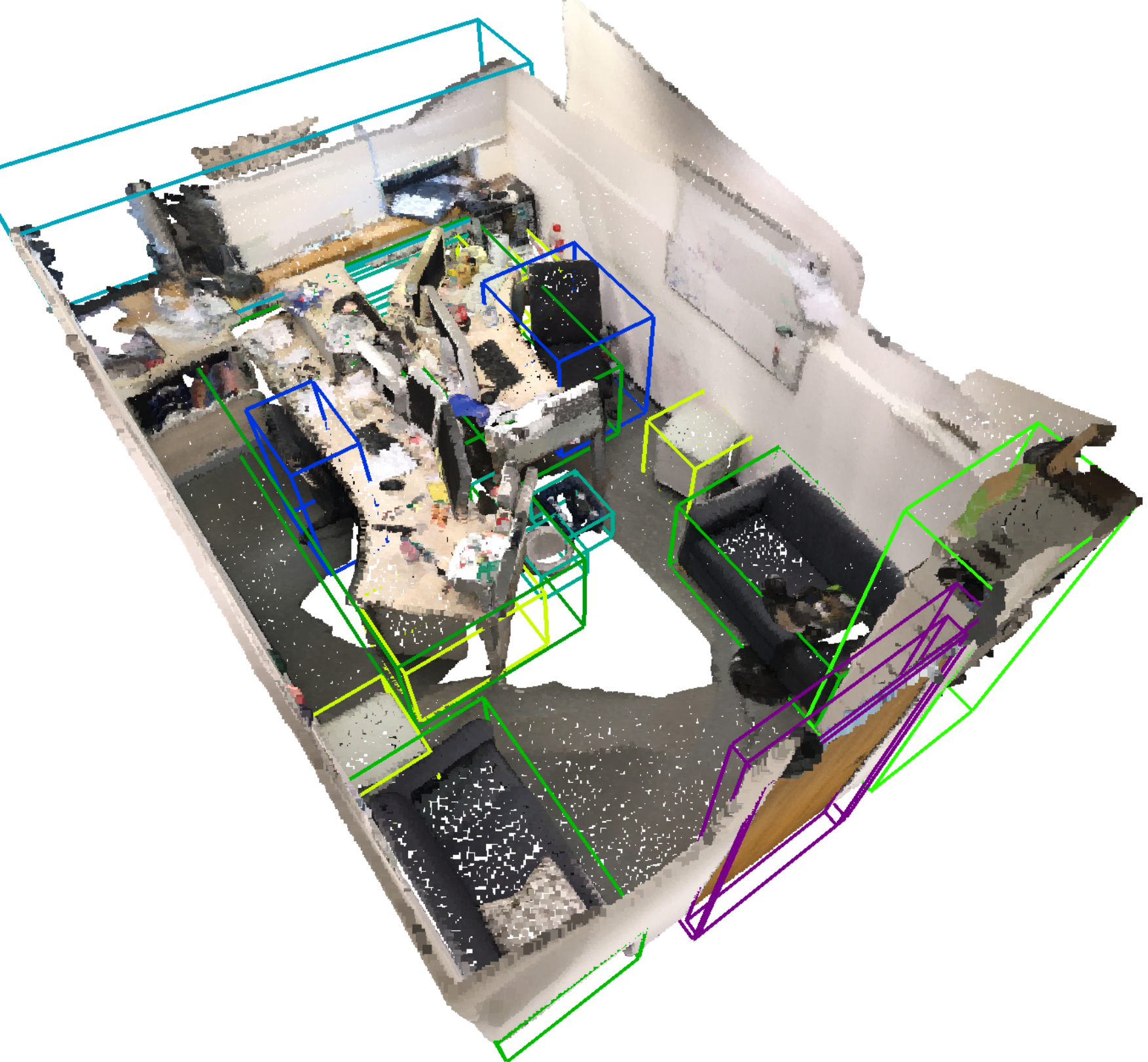}} &
   \makecell[l]{\includegraphics[width=0.24\linewidth]{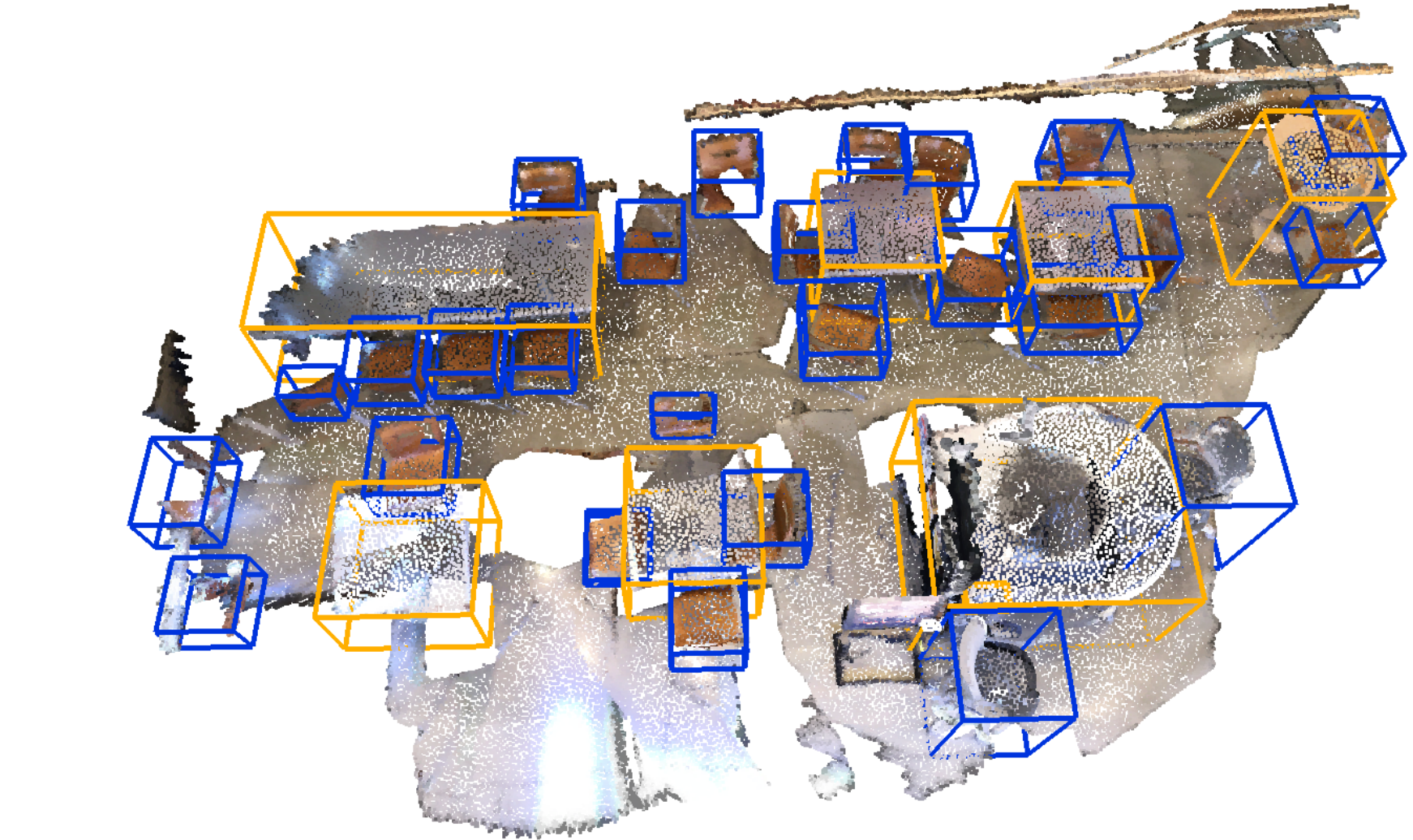}} &
   \makecell[l]{\includegraphics[width=0.24\linewidth]{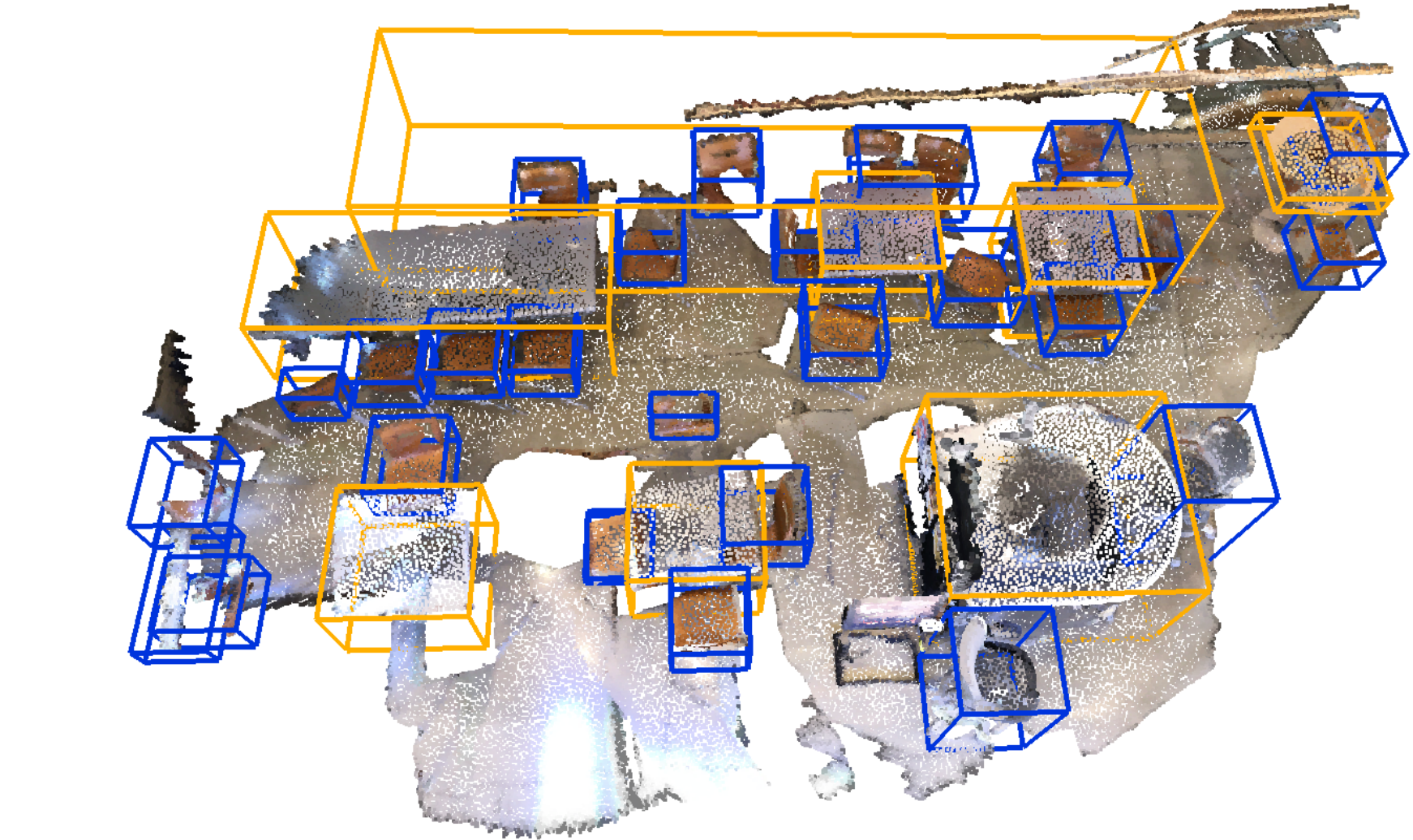}} \\
   \multicolumn{4}{c}{a) ScanNet} \\
   \makecell[l]{\includegraphics[width=0.24\linewidth]{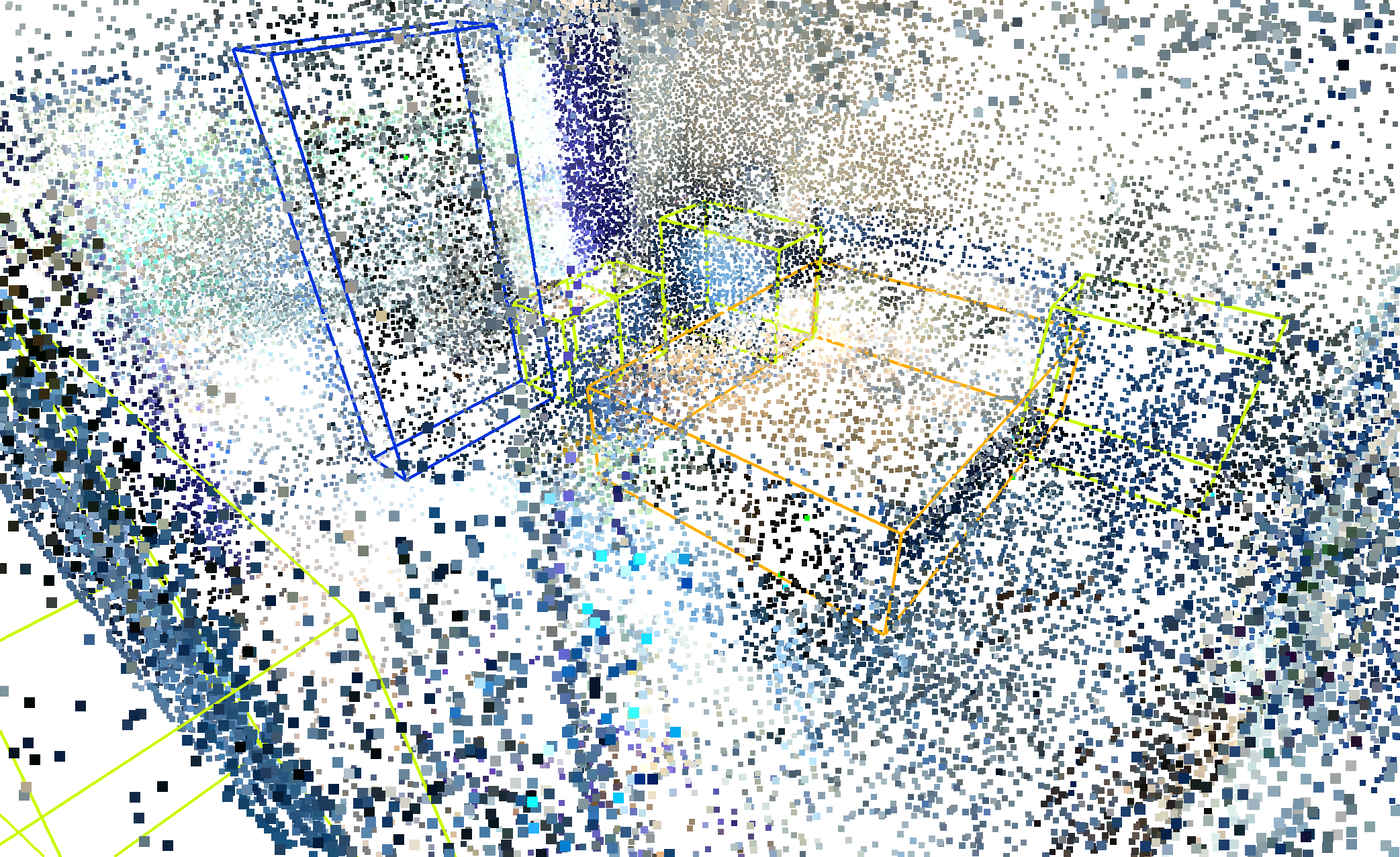}} &
   \makecell[l]{\includegraphics[width=0.24\linewidth]{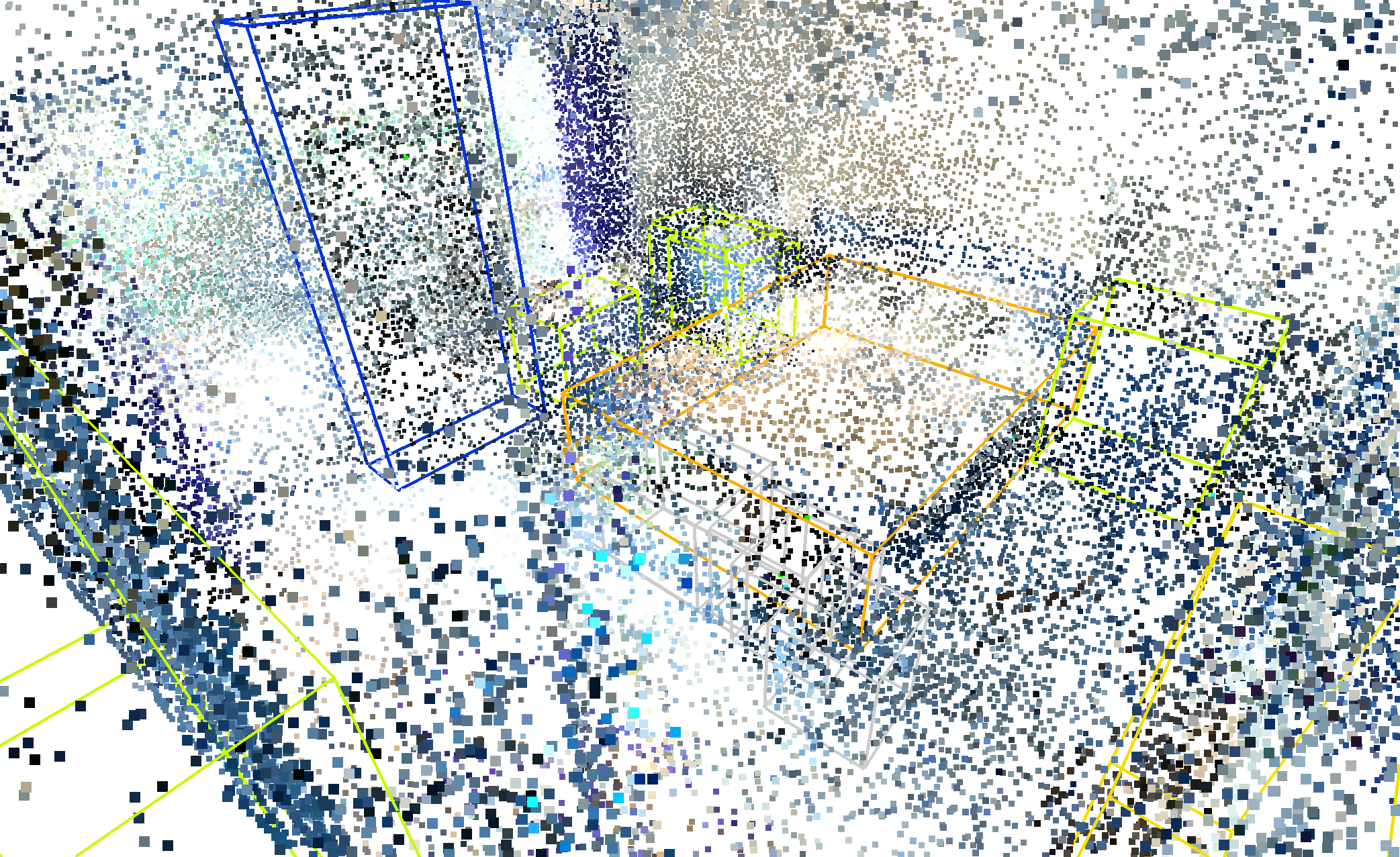}} &
   \makecell[l]{\includegraphics[width=0.24\linewidth]{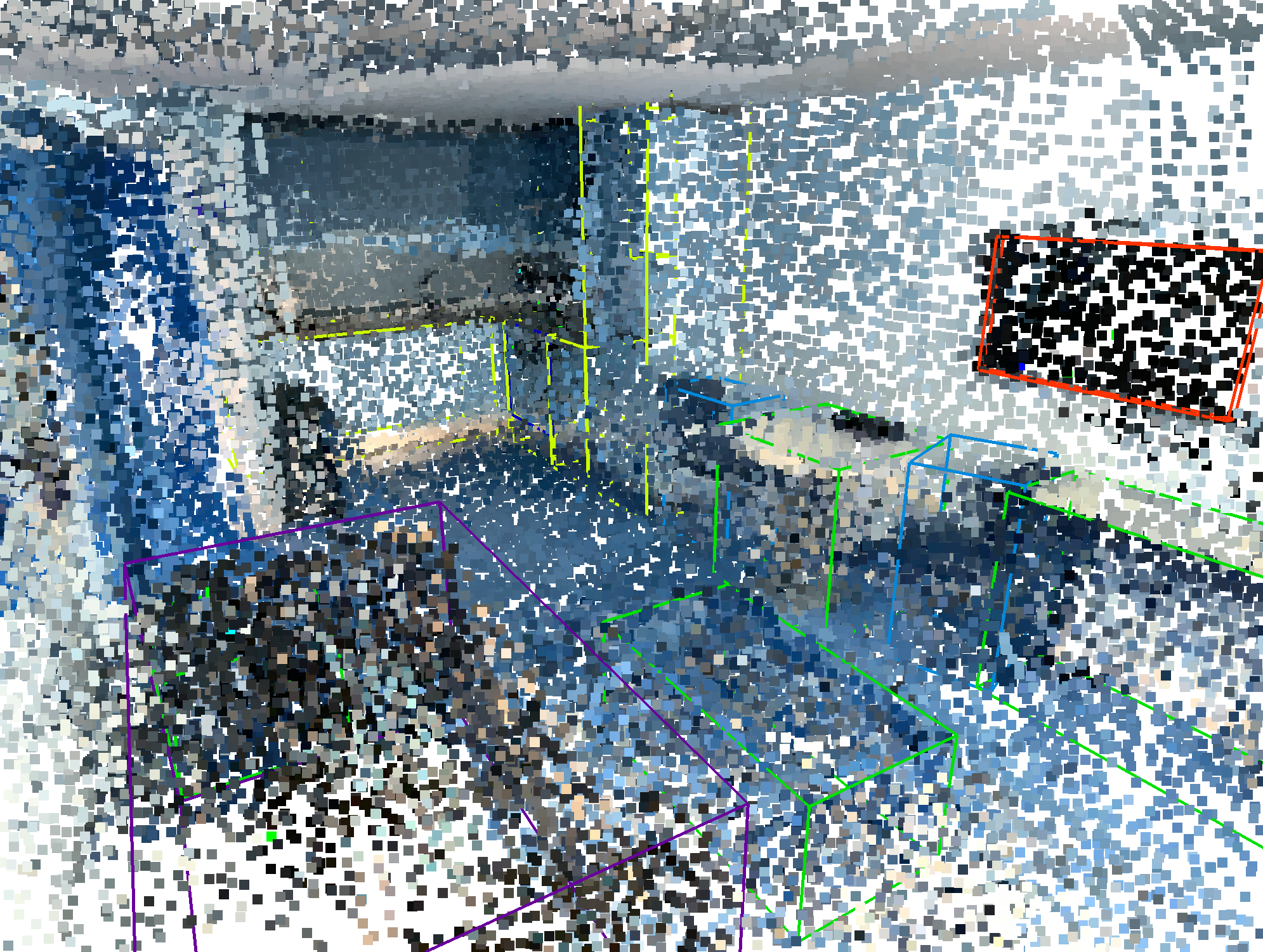}} &
   \makecell[l]{\includegraphics[width=0.24\linewidth]{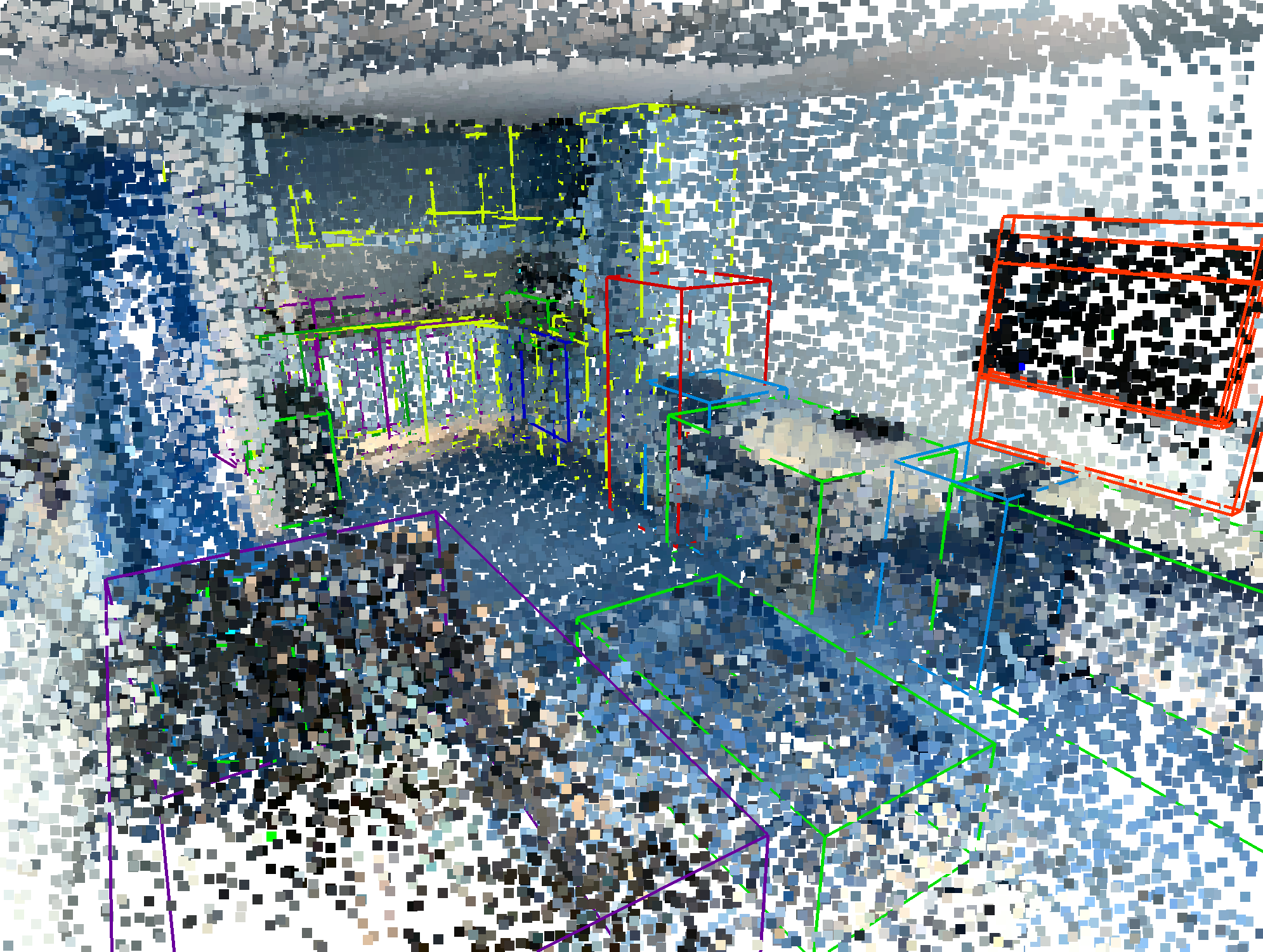}} \\
   \multicolumn{4}{c}{b) ARKitScenes} \\
   \makecell[l]{\includegraphics[width=0.24\linewidth]{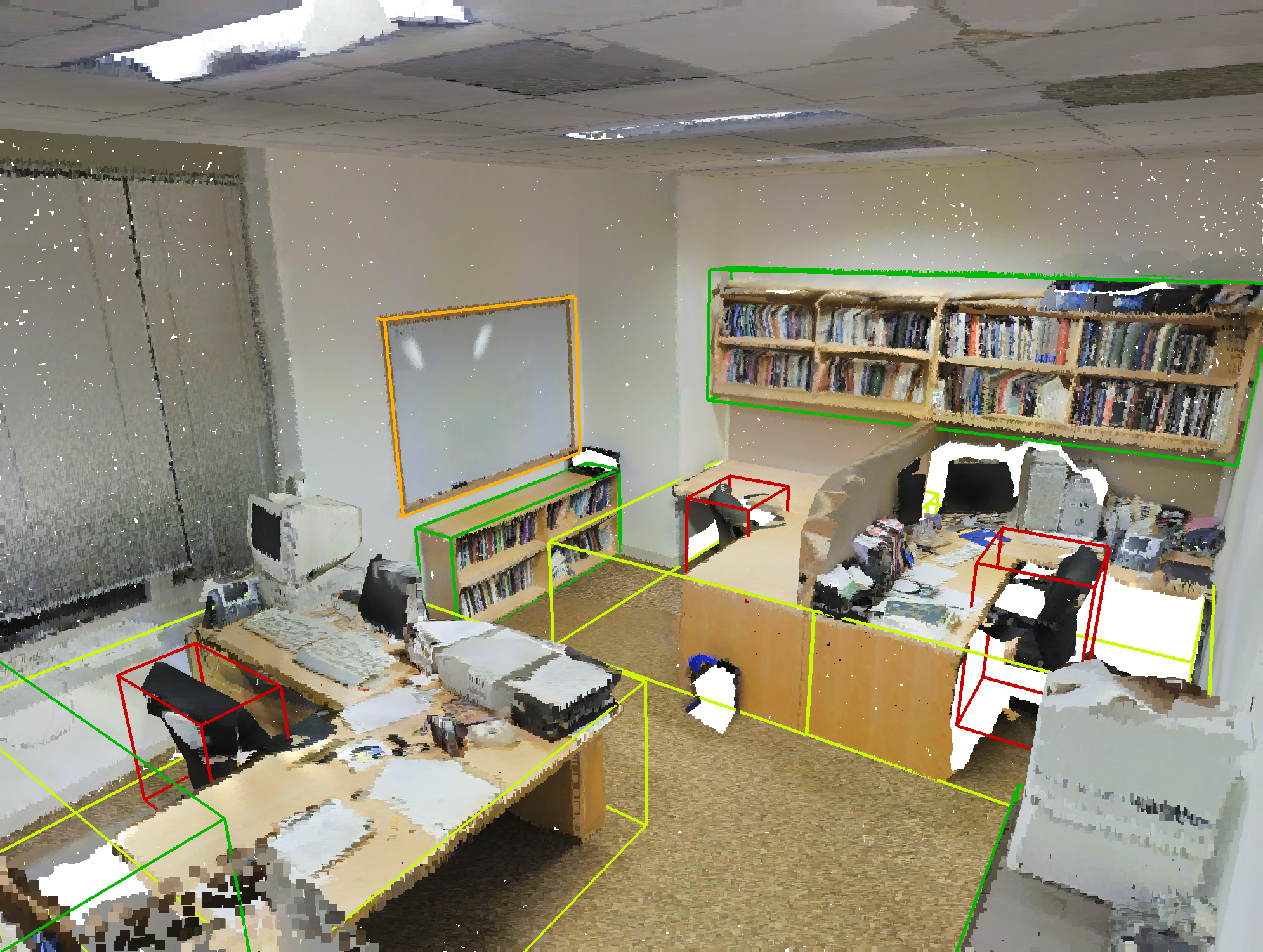}} &
   \makecell[l]{\includegraphics[width=0.24\linewidth]{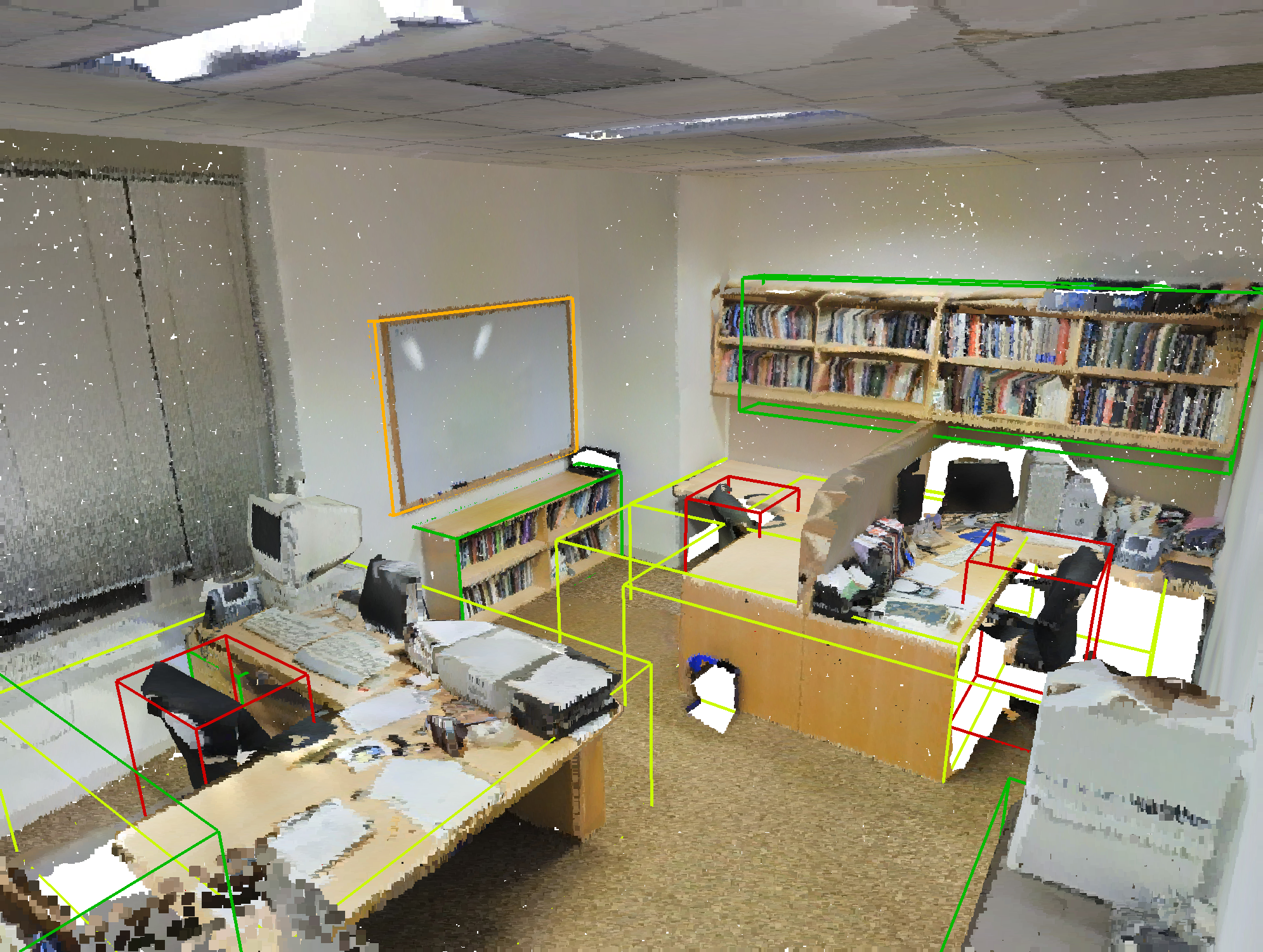}} &
   \makecell[l]{\includegraphics[width=0.24\linewidth]{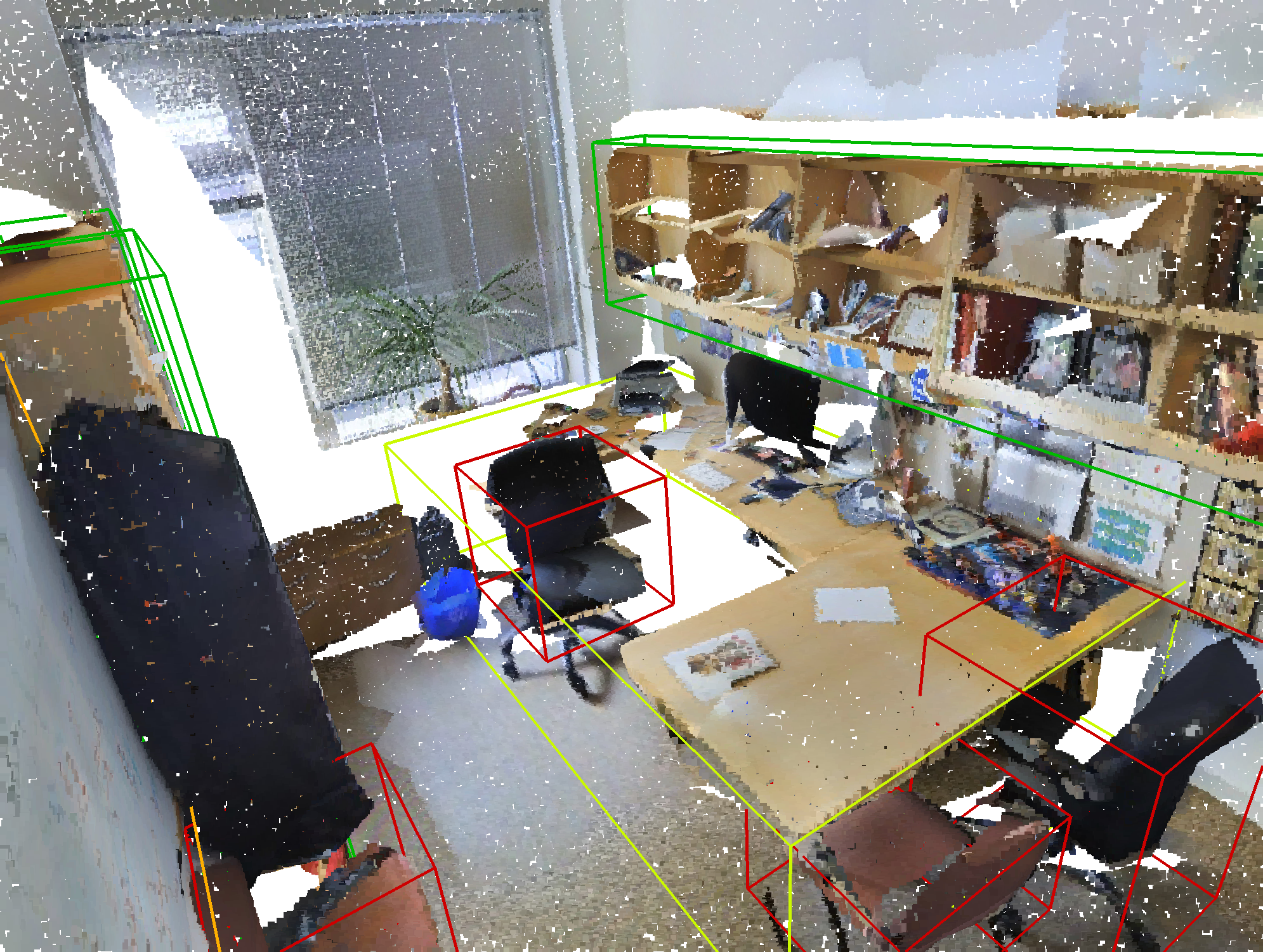}} &
   \makecell[l]{\includegraphics[width=0.24\linewidth]{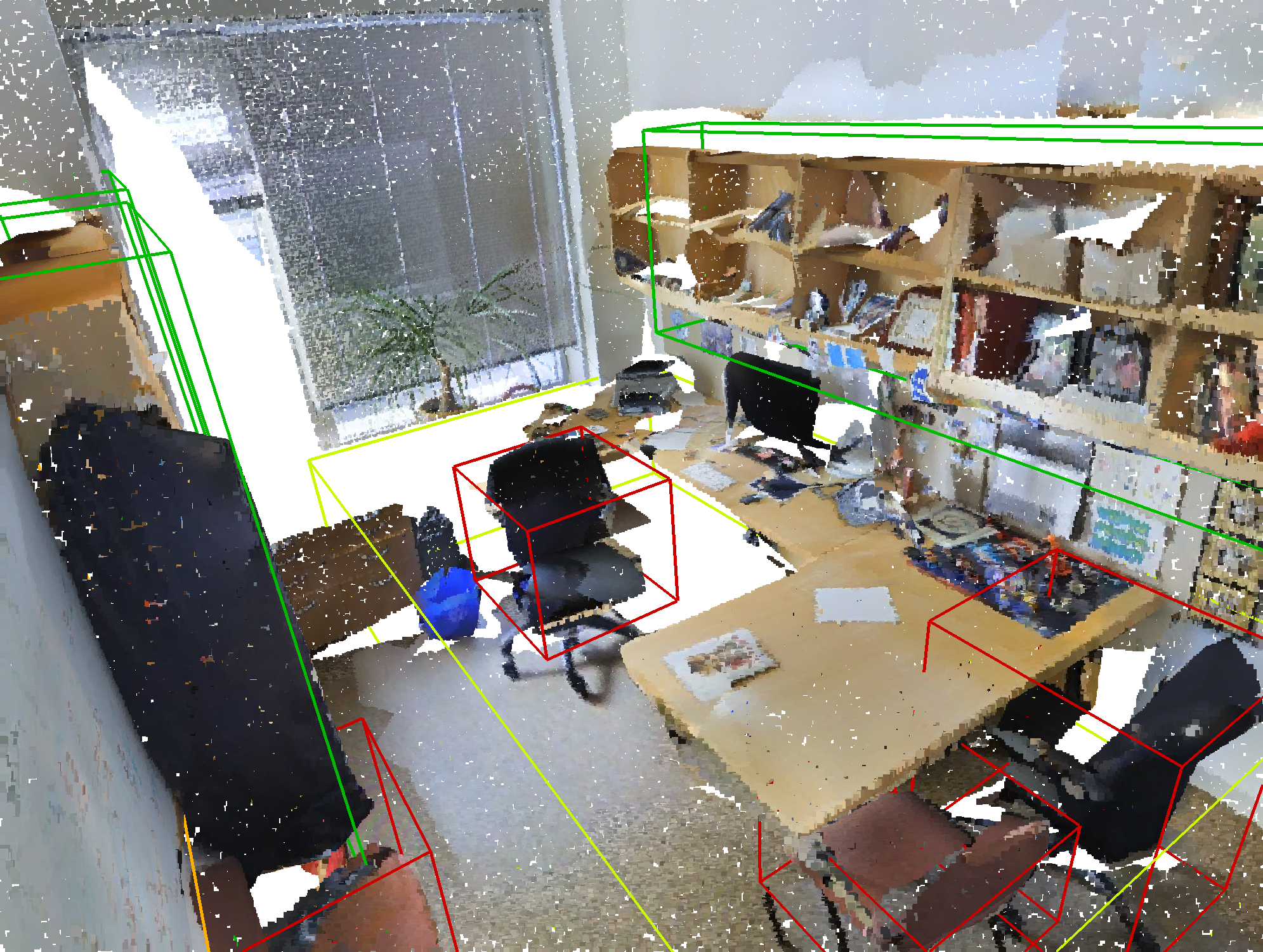}} \\
   \multicolumn{4}{c}{c) S3DIS} \\
   \makecell[l]{\includegraphics[width=0.24\linewidth]{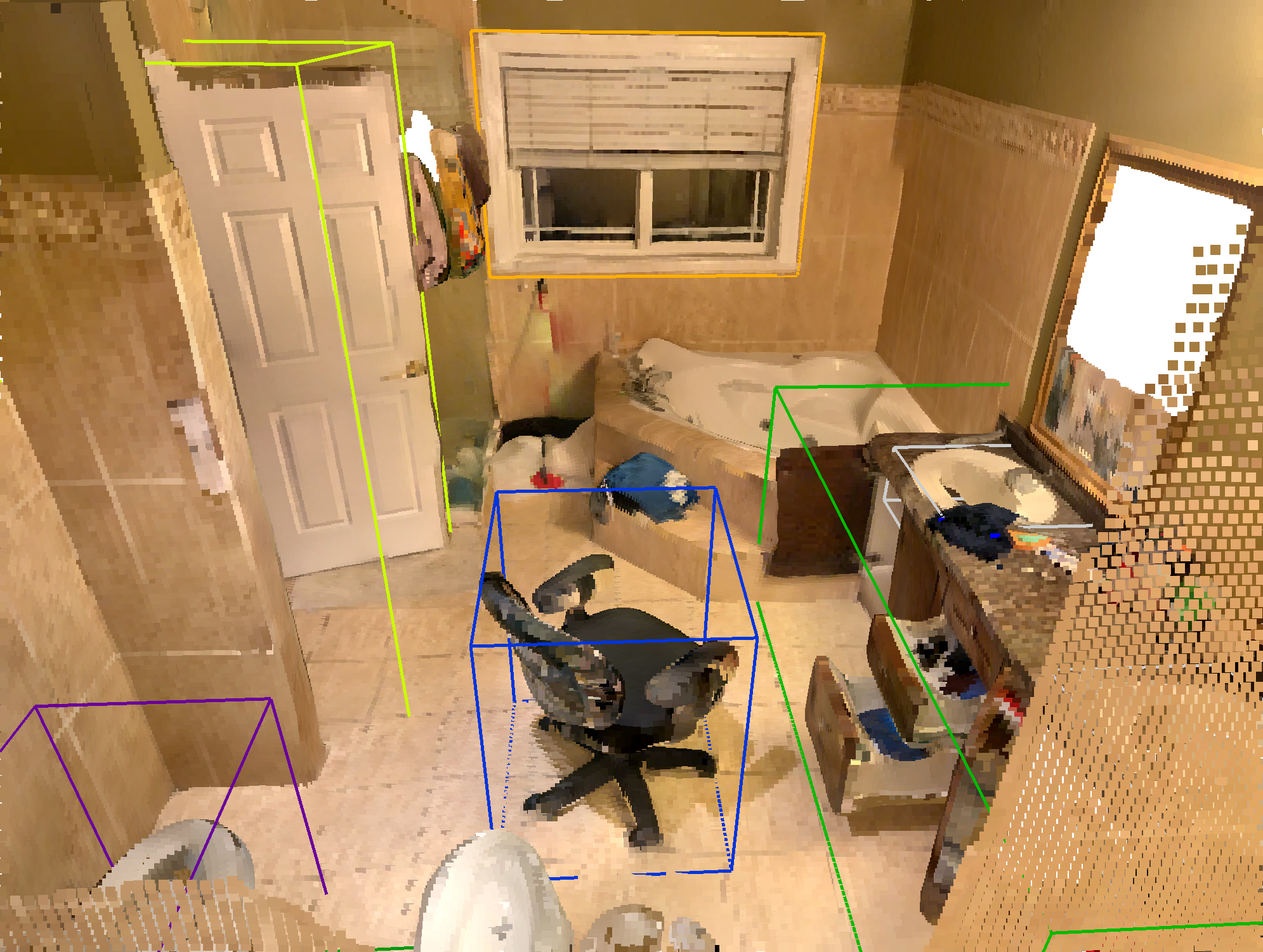}} &
   \makecell[l]{\includegraphics[width=0.24\linewidth]{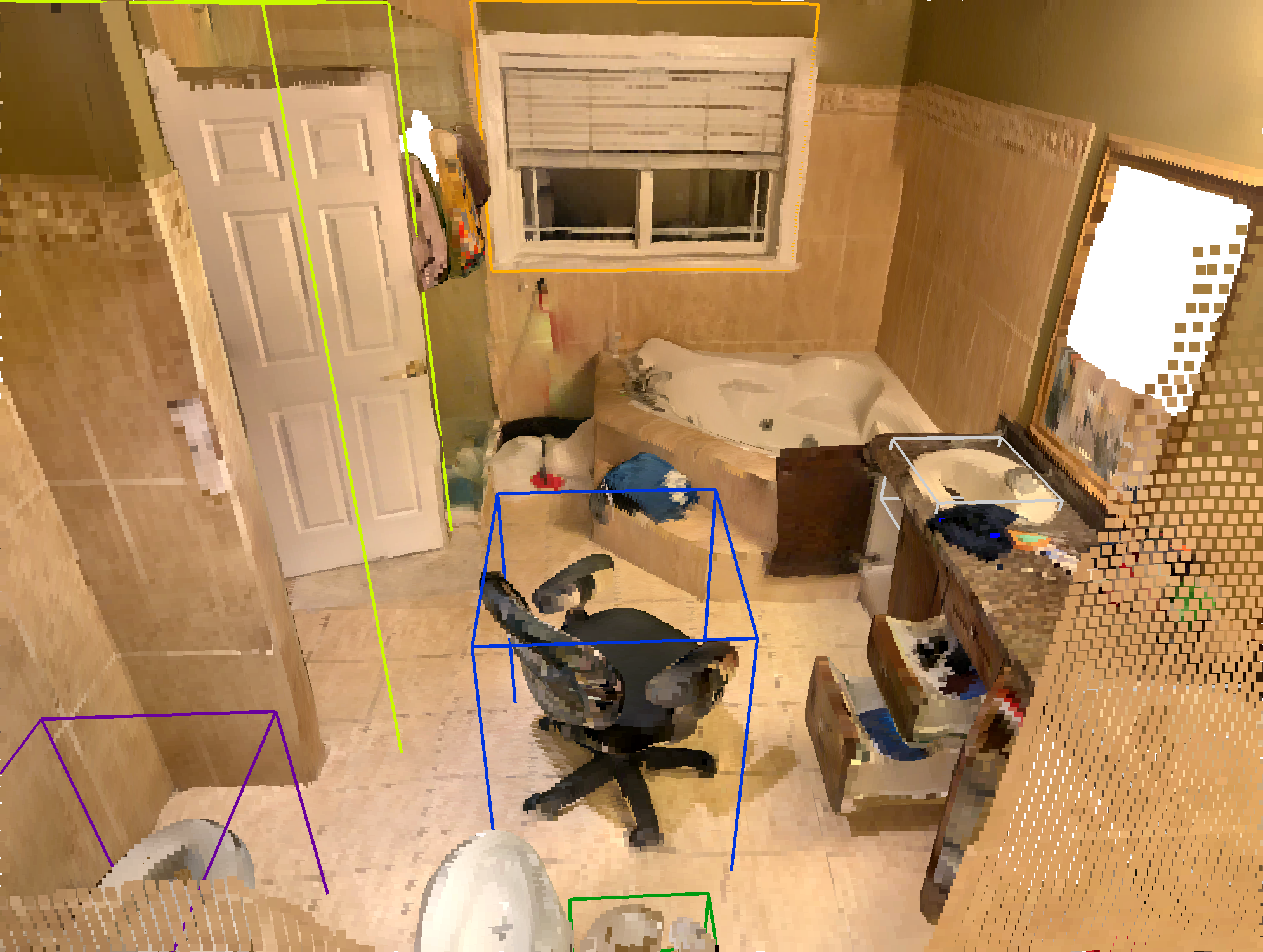}} &
   \makecell[l]{\includegraphics[width=0.24\linewidth]{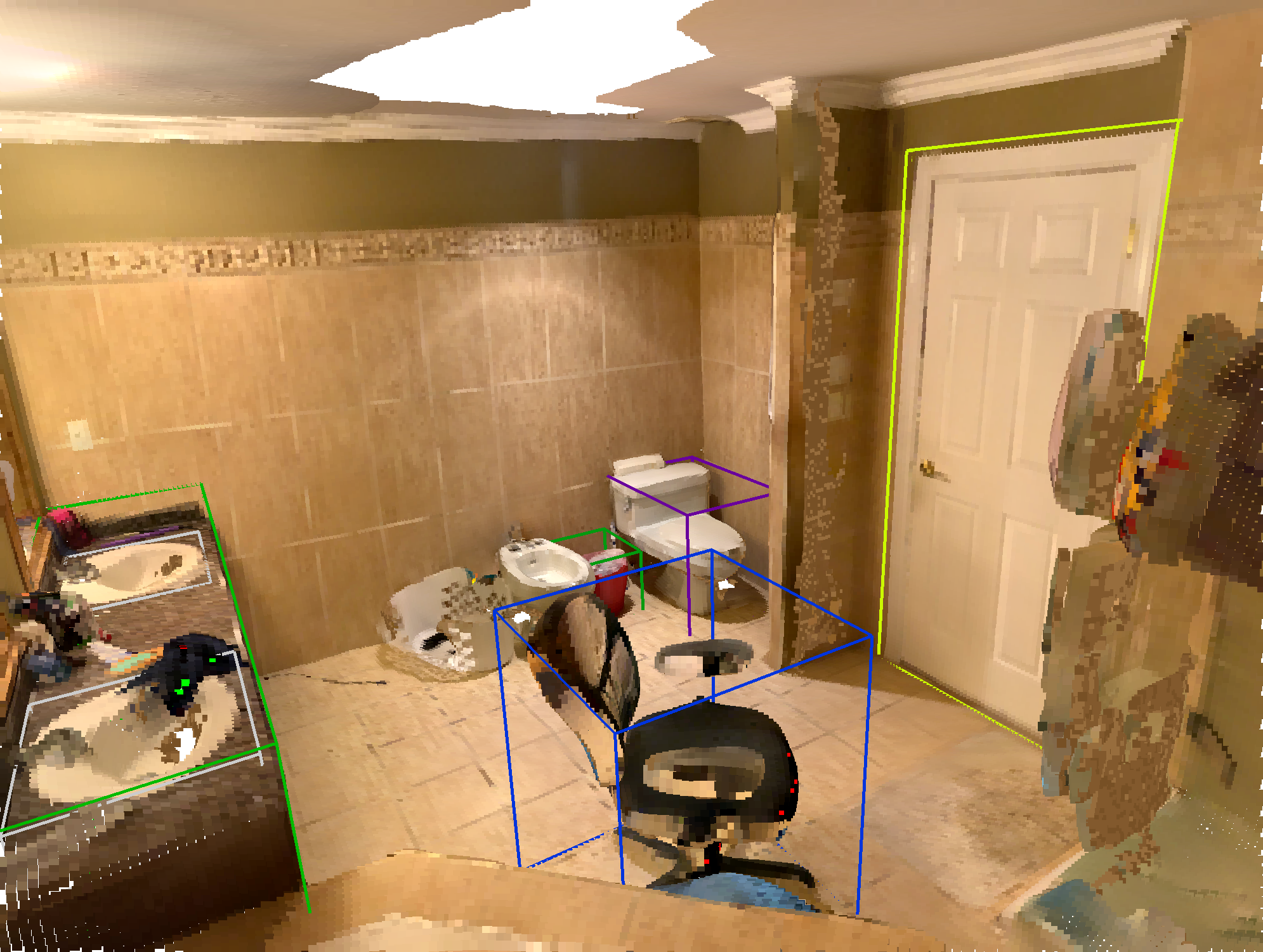}} &
   \makecell[l]{\includegraphics[width=0.24\linewidth]{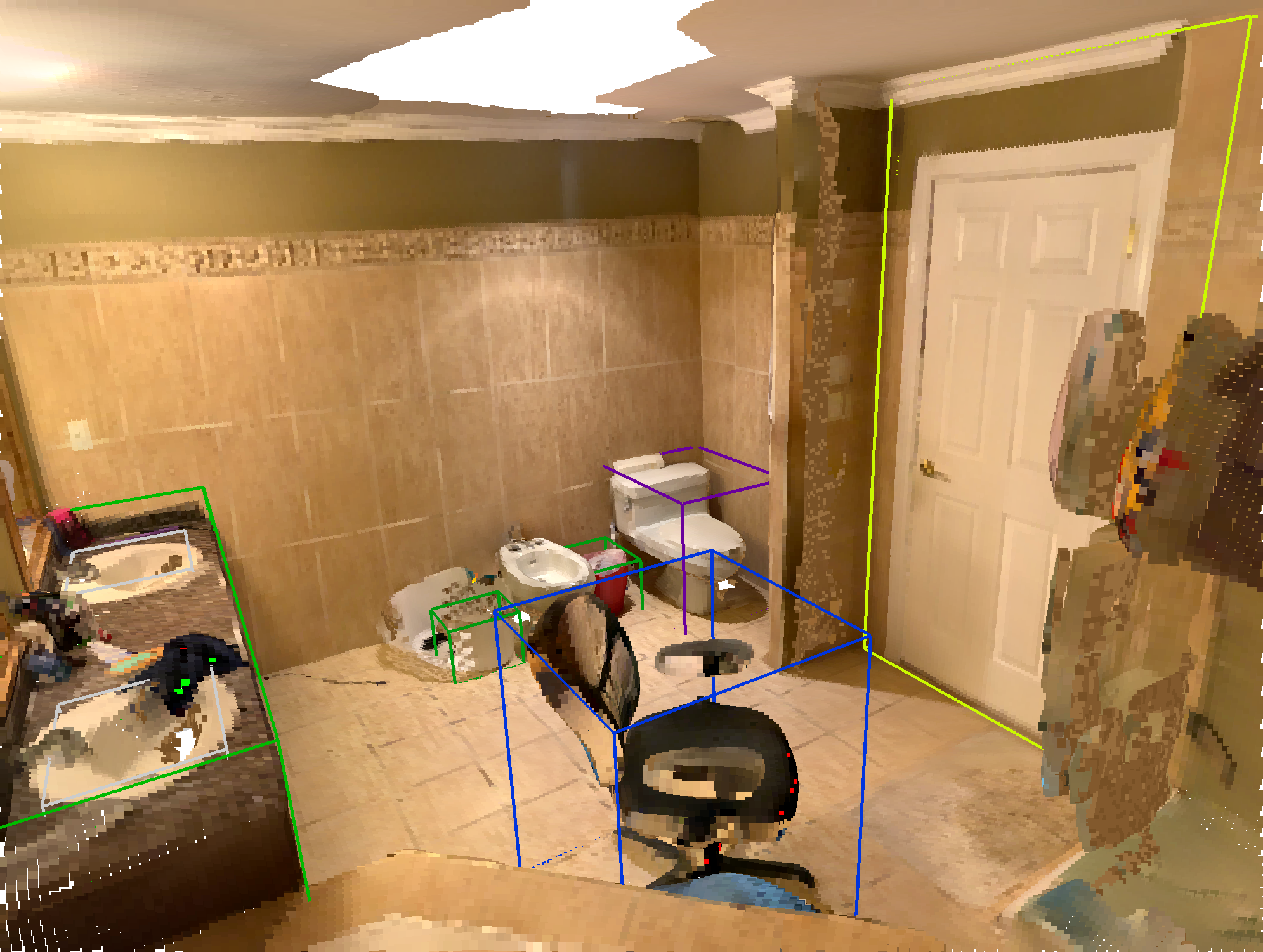}} \\
   \multicolumn{4}{c}{d) MultiScan} \\
   \makecell[l]{\includegraphics[width=0.24\linewidth]{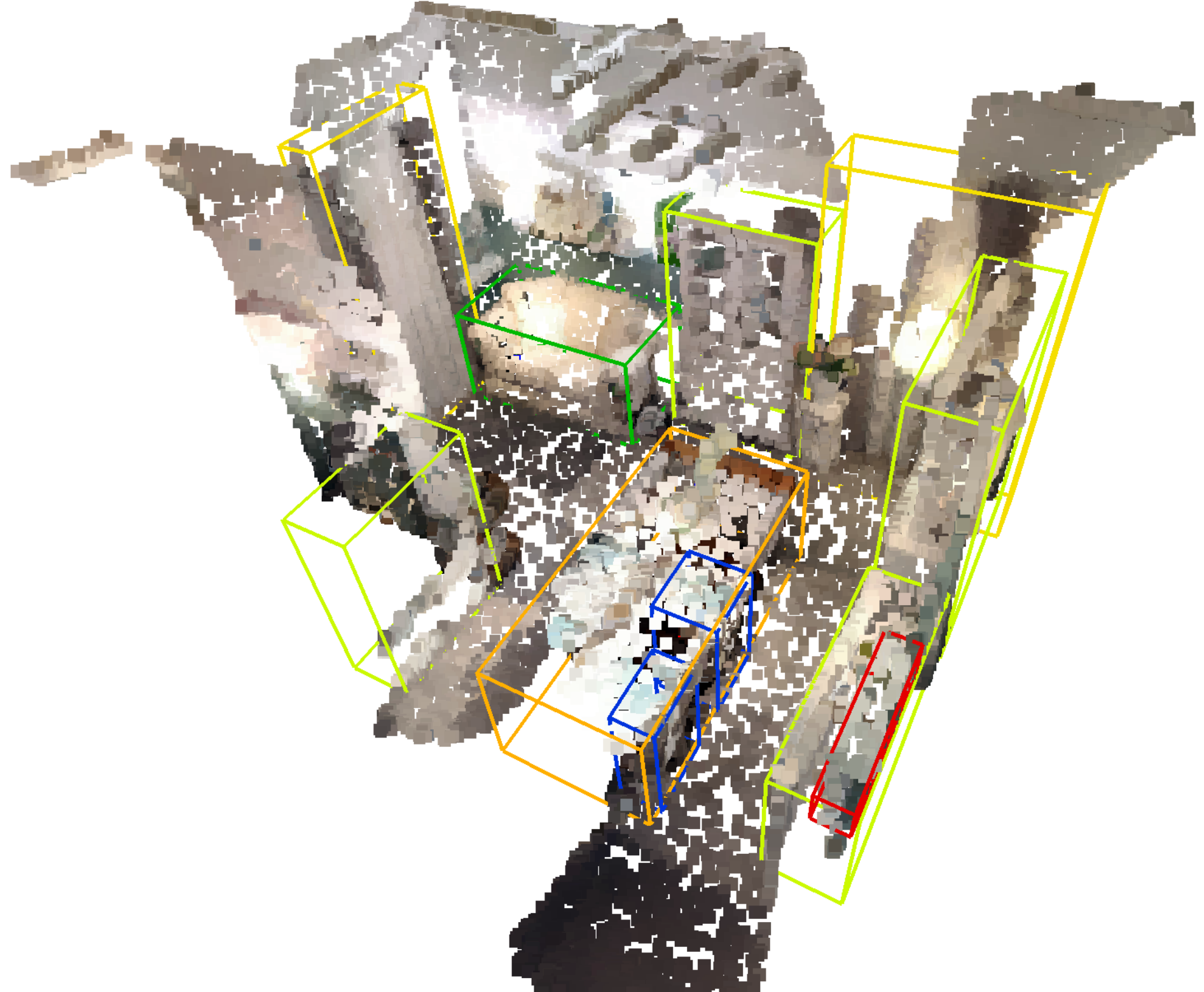}} &
   \makecell[l]{\includegraphics[width=0.24\linewidth]{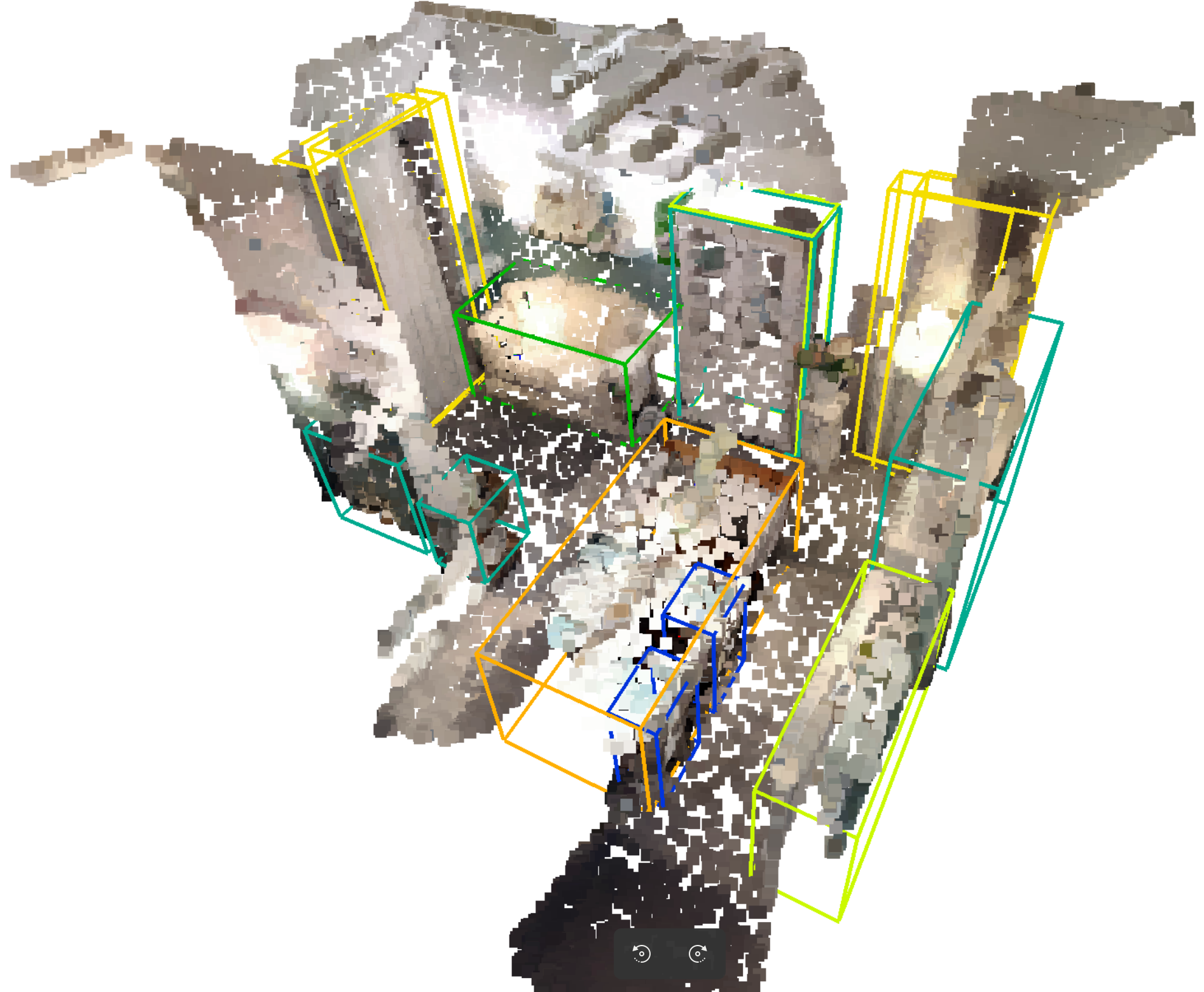}} &
   \makecell[l]{\includegraphics[width=0.24\linewidth]{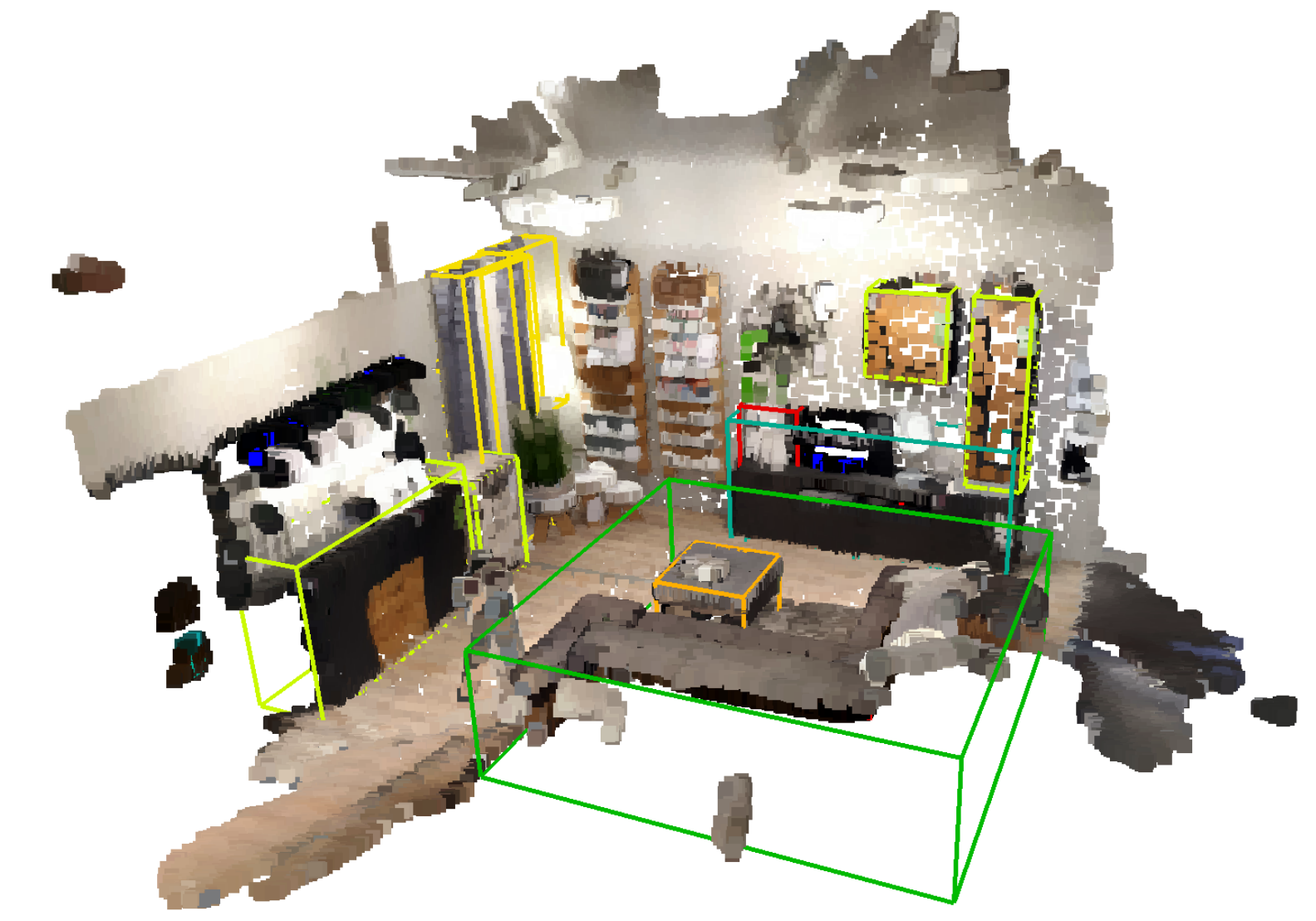}} &
   \makecell[l]{\includegraphics[width=0.24\linewidth]{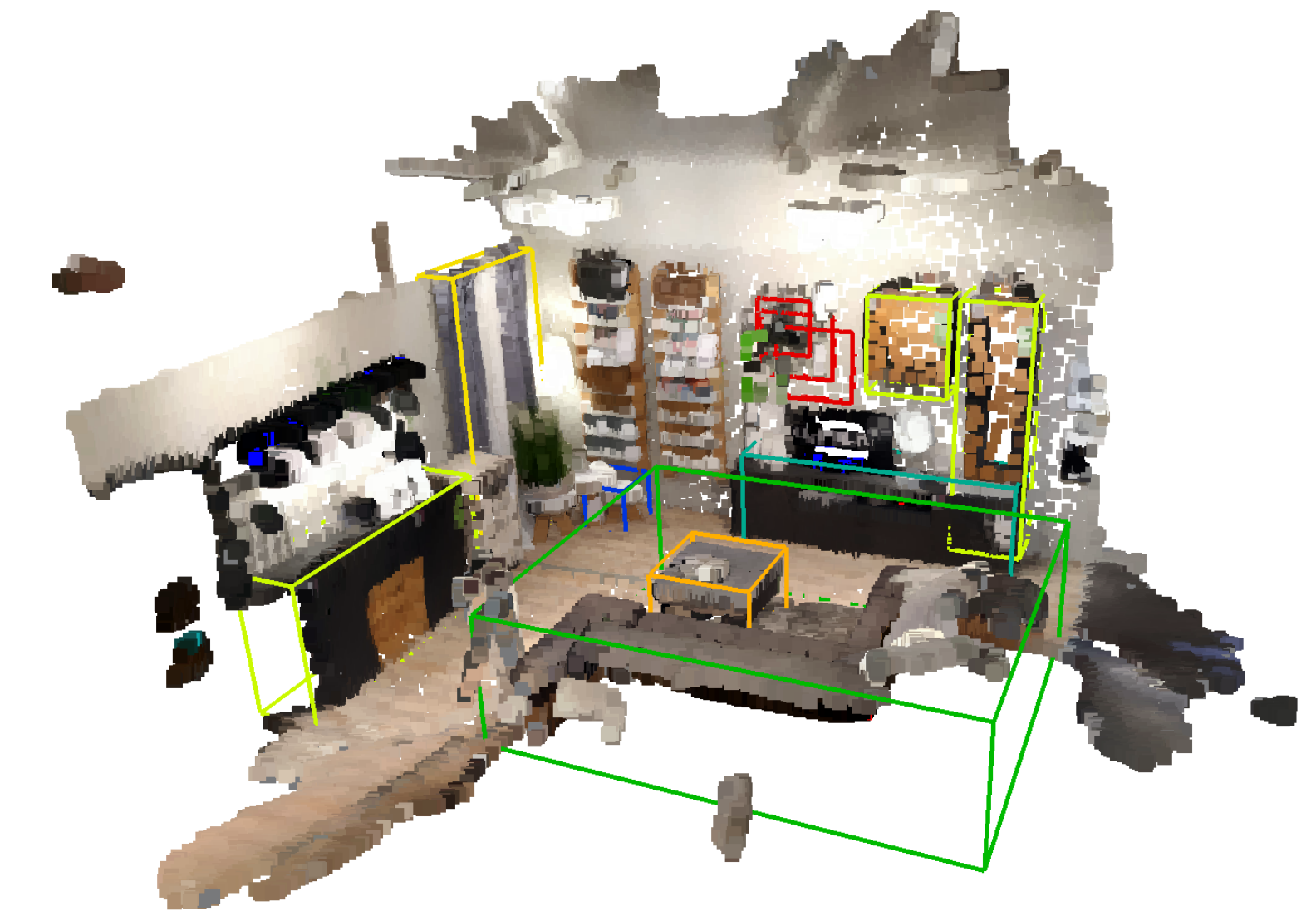}} \\
   \multicolumn{4}{c}{e) 3RScan} \\
   \makecell[l]{\includegraphics[width=0.24\linewidth]{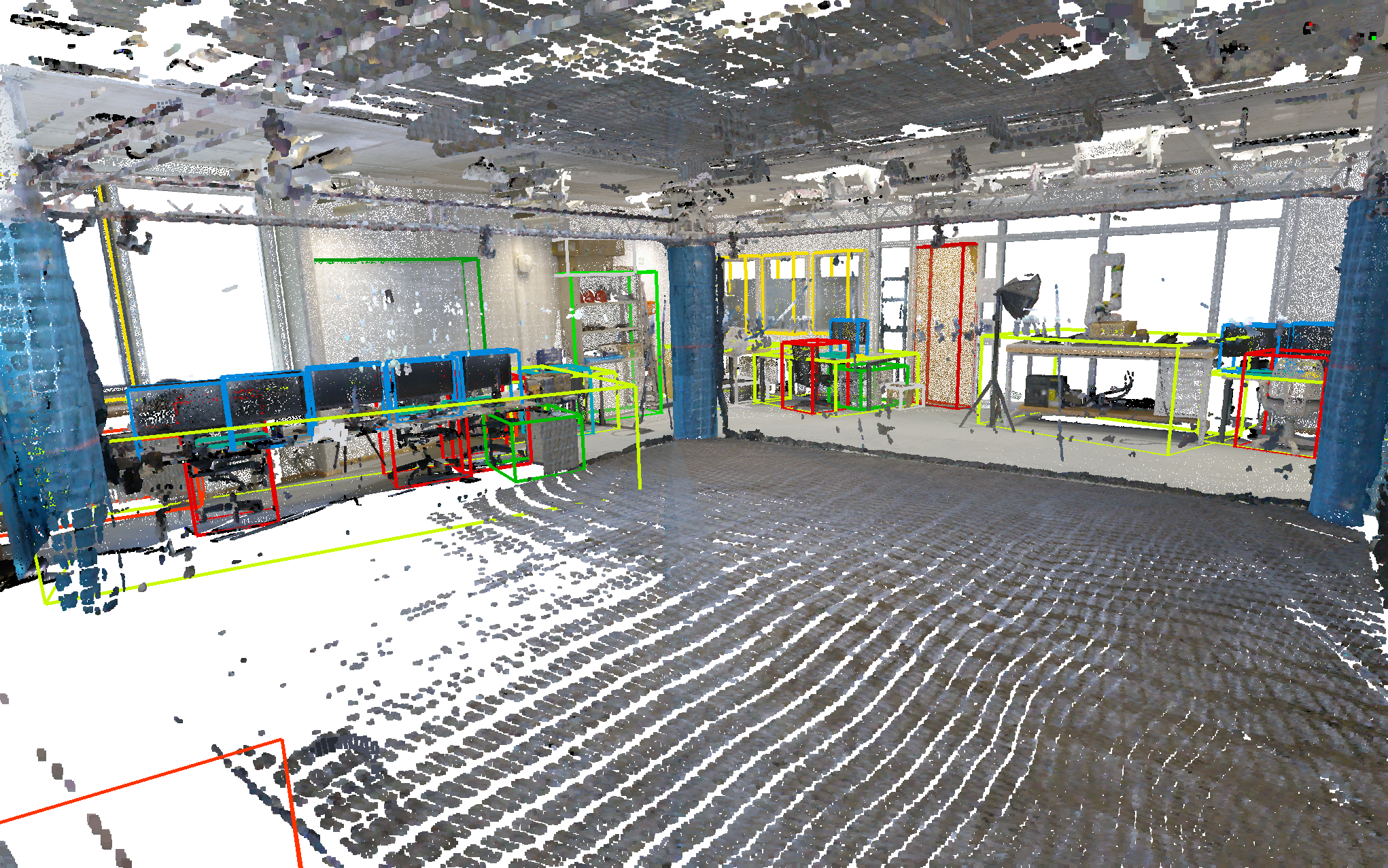}} &
   \makecell[l]{\includegraphics[width=0.24\linewidth]{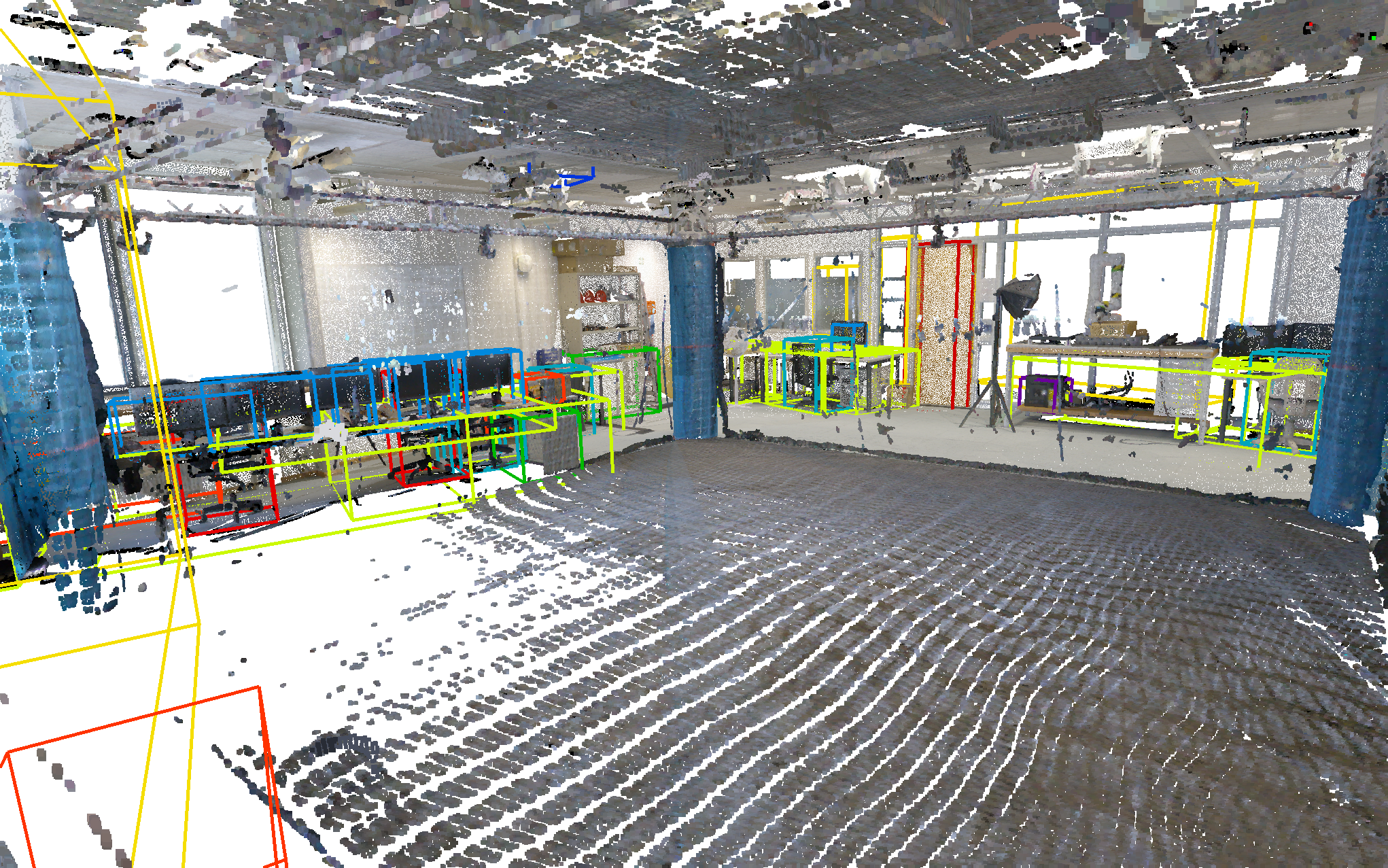}} &
   \makecell[l]{\includegraphics[width=0.24\linewidth]{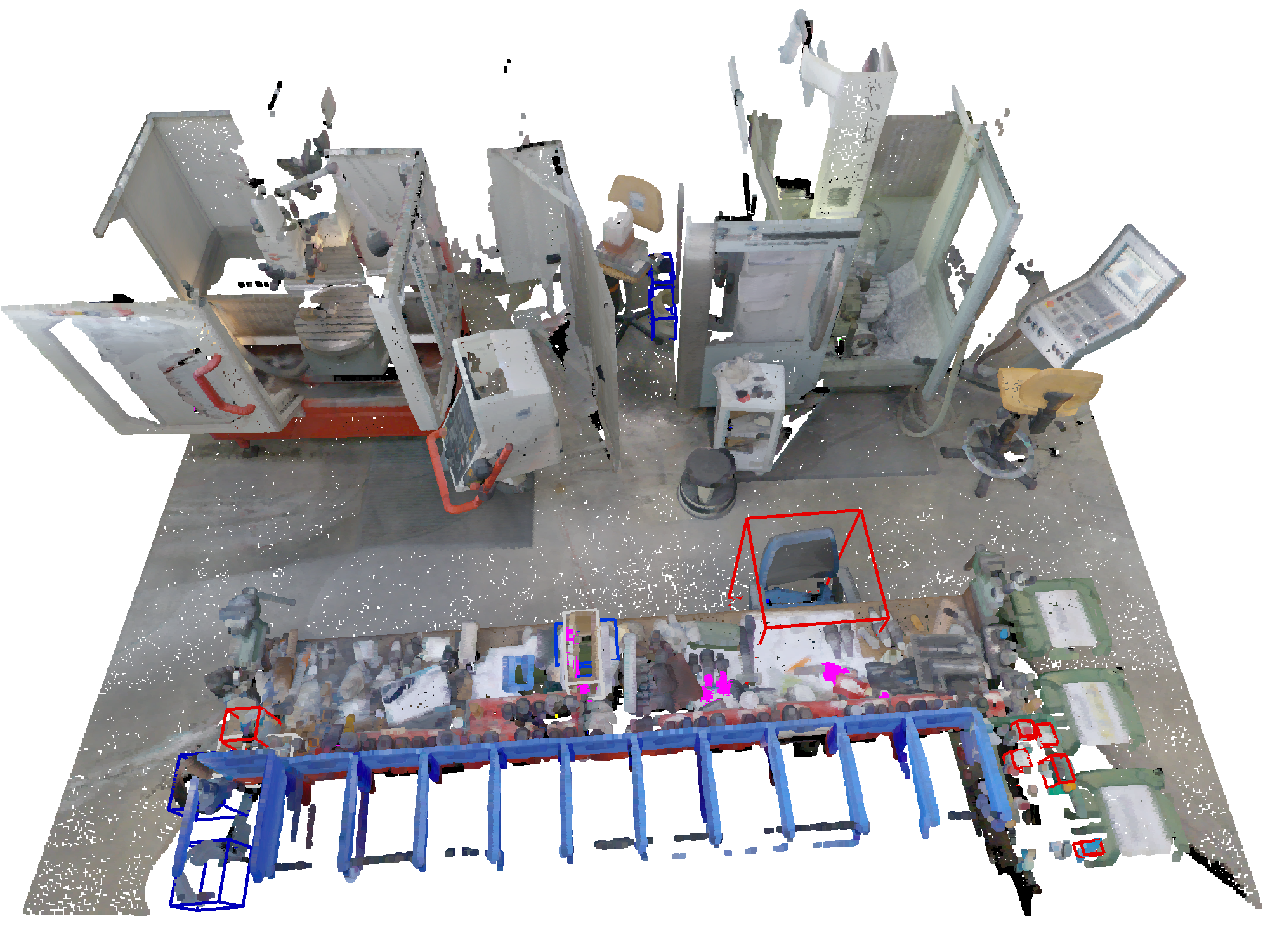}} &
   \makecell[l]{\includegraphics[width=0.24\linewidth]{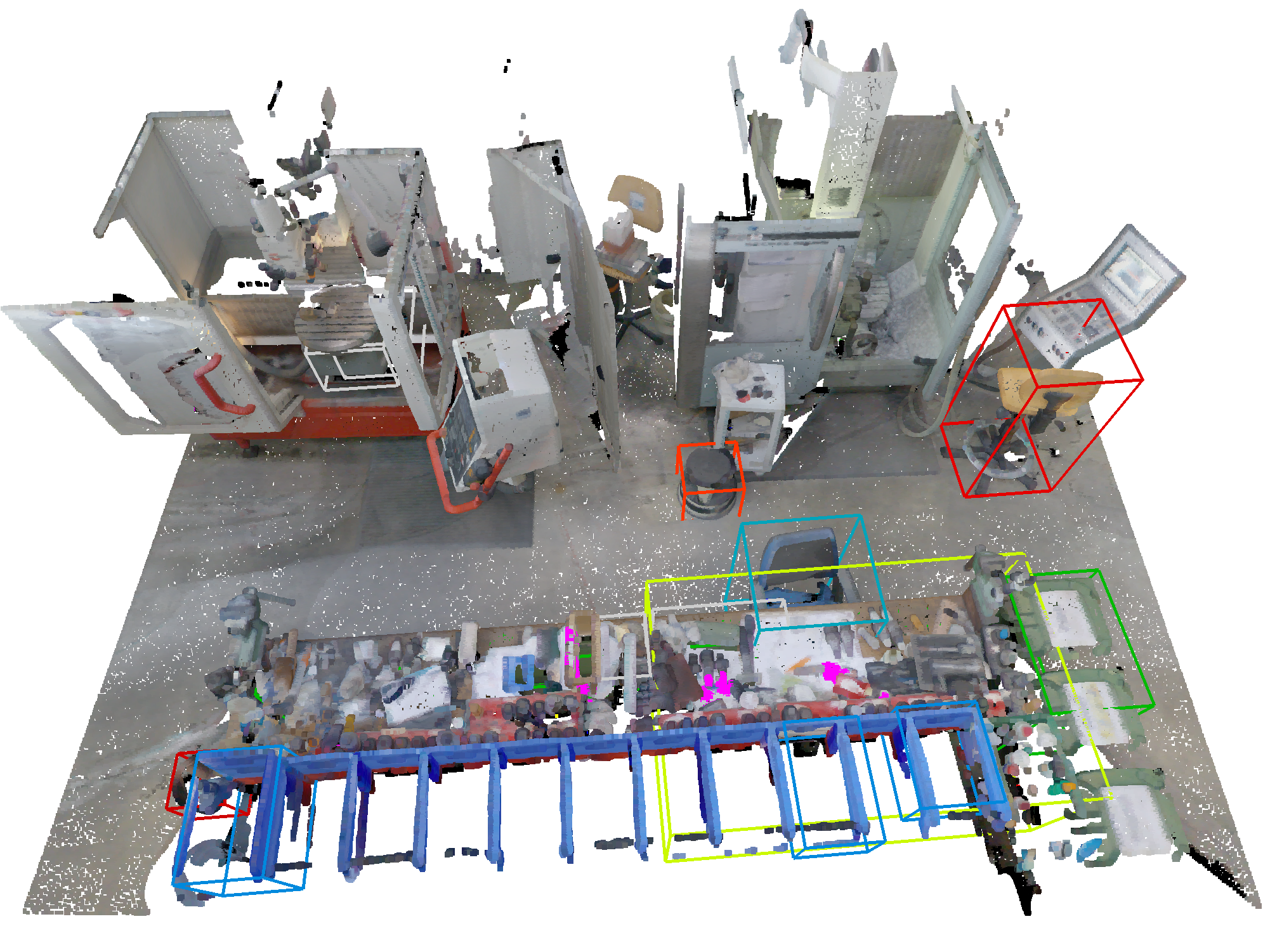}} \\
   \multicolumn{4}{c}{f) ScanNet++} \\
\end{tabular}
\caption{UniDet3D predictions compared to ground truth on six datasets.}
\label{fig:visualization}
\end{table*}

\clearpage
\bibliography{aaai25}

\begin{thebibliography}{54}
\providecommand{\natexlab}[1]{#1}

\bibitem[{Armeni et~al.(2016)Armeni, Sener, Zamir, Jiang, Brilakis, Fischer, and Savarese}]{armeni2016s3dis}
Armeni, I.; Sener, O.; Zamir, A.~R.; Jiang, H.; Brilakis, I.; Fischer, M.; and Savarese, S. 2016.
\newblock 3D Semantic Parsing of Large-Scale Indoor Spaces.
\newblock In \emph{IEEE Conference on Computer Vision and Pattern Recognition (CVPR)}, 1534--1543.

\bibitem[{Baruch et~al.(2021)Baruch, Chen, Dehghan, Dimry, Feigin, Fu, Gebauer, Joffe, Kurz, Schwartz et~al.}]{baruch2021arkitscenes}
Baruch, G.; Chen, Z.; Dehghan, A.; Dimry, T.; Feigin, Y.; Fu, P.; Gebauer, T.; Joffe, B.; Kurz, D.; Schwartz, A.; et~al. 2021.
\newblock ARKitScenes--A Diverse Real-World Dataset For 3D Indoor Scene Understanding Using Mobile RGB-D Data.
\newblock \emph{arXiv preprint arXiv:2111.08897}.

\bibitem[{Brazil et~al.(2023)Brazil, Kumar, Straub, Ravi, Johnson, and Gkioxari}]{brazil2023omni3d}
Brazil, G.; Kumar, A.; Straub, J.; Ravi, N.; Johnson, J.; and Gkioxari, G. 2023.
\newblock Omni3d: A large benchmark and model for 3d object detection in the wild.
\newblock In \emph{Proceedings of the IEEE/CVF conference on computer vision and pattern recognition}, 13154--13164.

\bibitem[{Cai et~al.(2022)Cai, Zhang, Zhu, Zhang, Li, and Xue}]{cai2022bigdetection}
Cai, L.; Zhang, Z.; Zhu, Y.; Zhang, L.; Li, M.; and Xue, X. 2022.
\newblock Bigdetection: A large-scale benchmark for improved object detector pre-training.
\newblock In \emph{Proceedings of the IEEE/CVF conference on computer vision and pattern recognition}, 4777--4787.

\bibitem[{Chen et~al.(2020)Chen, Lei, Song, Ying, Chen, and Wu}]{chen2020hgnet}
Chen, J.; Lei, B.; Song, Q.; Ying, H.; Chen, D.~Z.; and Wu, J. 2020.
\newblock A hierarchical graph network for 3D object detection on point clouds.
\newblock In \emph{Proceedings of the IEEE/CVF conference on computer vision and pattern recognition}, 392--401.

\bibitem[{Chen et~al.(2021)Chen, Fang, Zhang, Liu, and Wang}]{chen2021hais}
Chen, S.; Fang, J.; Zhang, Q.; Liu, W.; and Wang, X. 2021.
\newblock Hierarchical aggregation for 3d instance segmentation.
\newblock In \emph{Proceedings of the IEEE/CVF International Conference on Computer Vision}, 15467--15476.

\bibitem[{Cheng et~al.(2021)Cheng, Sheng, Shi, Yang, and Xu}]{cheng2021brnet}
Cheng, B.; Sheng, L.; Shi, S.; Yang, M.; and Xu, D. 2021.
\newblock Back-tracing Representative Points for Voting-based 3D Object Detection in Point Clouds.
\newblock In \emph{Proceedings of the IEEE/CVF Conference on Computer Vision and Pattern Recognition}, 8963--8972.

\bibitem[{Choy, Gwak, and Savarese(2019)}]{choy2019minkowski}
Choy, C.; Gwak, J.; and Savarese, S. 2019.
\newblock 4D Spatio-Temporal ConvNets: Minkowski Convolutional Neural Networks.
\newblock In \emph{IEEE Conference on Computer Vision and Pattern Recognition (CVPR)}, 3075--3084.

\bibitem[{Contributors(2020)}]{2020mmdetection3d}
Contributors, M. 2020.
\newblock {MMDetection3D: OpenMMLab} next-generation platform for general {3D} object detection.
\newblock \url{https://github.com/open-mmlab/mmdetection3d}.

\bibitem[{Contributors(2022)}]{spconv2022}
Contributors, S. 2022.
\newblock Spconv: Spatially Sparse Convolution Library.
\newblock \url{https://github.com/traveller59/spconv}.

\bibitem[{Dai et~al.(2017)Dai, Chang, Savva, Halber, Funkhouser, and Nie{\ss}ner}]{dai2017scannet}
Dai, A.; Chang, A.~X.; Savva, M.; Halber, M.; Funkhouser, T.; and Nie{\ss}ner, M. 2017.
\newblock Scannet: Richly-annotated 3d reconstructions of indoor scenes.
\newblock In \emph{Proceedings of the IEEE conference on computer vision and pattern recognition}, 5828--5839.

\bibitem[{Engelmann et~al.(2020)Engelmann, Bokeloh, Fathi, Leibe, and Nie{\ss}ner}]{engelmann20203d-mpa}
Engelmann, F.; Bokeloh, M.; Fathi, A.; Leibe, B.; and Nie{\ss}ner, M. 2020.
\newblock 3d-mpa: Multi-proposal aggregation for 3d semantic instance segmentation.
\newblock In \emph{Proceedings of the IEEE/CVF conference on computer vision and pattern recognition}, 9031--9040.

\bibitem[{Gwak, Choy, and Savarese(2020)}]{gwak2020gsdn}
Gwak, J.; Choy, C.; and Savarese, S. 2020.
\newblock Generative sparse detection networks for 3d single-shot object detection.
\newblock In \emph{Computer Vision--ECCV 2020: 16th European Conference, Glasgow, UK, August 23--28, 2020, Proceedings, Part IV 16}, 297--313. Springer.

\bibitem[{He, Shen, and Van Den~Hengel(2021)}]{he2021dyco3d}
He, T.; Shen, C.; and Van Den~Hengel, A. 2021.
\newblock Dyco3d: Robust instance segmentation of 3d point clouds through dynamic convolution.
\newblock In \emph{Proceedings of the IEEE/CVF conference on computer vision and pattern recognition}, 354--363.

\bibitem[{Jia et~al.(2024)Jia, Chen, Yu, Wang, Niu, Liu, Li, and Huang}]{jia2024sceneverse}
Jia, B.; Chen, Y.; Yu, H.; Wang, Y.; Niu, X.; Liu, T.; Li, Q.; and Huang, S. 2024.
\newblock Sceneverse: Scaling 3d vision-language learning for grounded scene understanding.
\newblock \emph{arXiv preprint arXiv:2401.09340}.

\bibitem[{Jiang et~al.(2020)Jiang, Zhao, Shi, Liu, Fu, and Jia}]{jiang2020pointgroup}
Jiang, L.; Zhao, H.; Shi, S.; Liu, S.; Fu, C.-W.; and Jia, J. 2020.
\newblock Pointgroup: Dual-set point grouping for 3d instance segmentation.
\newblock In \emph{Proceedings of the IEEE/CVF conference on computer vision and Pattern recognition}, 4867--4876.

\bibitem[{Kolodiazhnyi et~al.(2024)Kolodiazhnyi, Vorontsova, Konushin, and Rukhovich}]{kolodiazhnyi2024oneformer3d}
Kolodiazhnyi, M.; Vorontsova, A.; Konushin, A.; and Rukhovich, D. 2024.
\newblock Oneformer3d: One transformer for unified point cloud segmentation.
\newblock In \emph{Proceedings of the IEEE/CVF Conference on Computer Vision and Pattern Recognition}, 20943--20953.

\bibitem[{Kuhn(1955)}]{kuhn1955hungarian}
Kuhn, H.~W. 1955.
\newblock The Hungarian method for the assignment problem.
\newblock \emph{Naval research logistics quarterly}, 2(1-2): 83--97.

\bibitem[{Landrieu and Simonovsky(2018)}]{landrieu2018superpoints}
Landrieu, L.; and Simonovsky, M. 2018.
\newblock Large-scale point cloud semantic segmentation with superpoint graphs.
\newblock In \emph{Proceedings of the IEEE conference on computer vision and pattern recognition}, 4558--4567.

\bibitem[{Li et~al.(2024)Li, Xu, Lim, and Zhao}]{li2024unimode}
Li, Z.; Xu, X.; Lim, S.; and Zhao, H. 2024.
\newblock UniMODE: Unified Monocular 3D Object Detection.
\newblock In \emph{Proceedings of the IEEE/CVF Conference on Computer Vision and Pattern Recognition}, 16561--16570.

\bibitem[{Liang et~al.(2021)Liang, Li, Xu, Tan, and Jia}]{liang2021sstnet}
Liang, Z.; Li, Z.; Xu, S.; Tan, M.; and Jia, K. 2021.
\newblock Instance segmentation in 3d scenes using semantic superpoint tree networks.
\newblock In \emph{Proceedings of the IEEE/CVF international conference on computer vision}, 2783--2792.

\bibitem[{Liu et~al.(2021)Liu, Zhang, Cao, Hu, and Tong}]{liu2021group-free}
Liu, Z.; Zhang, Z.; Cao, Y.; Hu, H.; and Tong, X. 2021.
\newblock Group-Free 3D Object Detection via Transformers.
\newblock In \emph{Proceedings of the IEEE/CVF International Conference on Computer Vision (ICCV)}, 2949--2958.

\bibitem[{Lu et~al.(2023{\natexlab{a}})Lu, Deng, Wang, He, and Zhang}]{lu2023queryformer}
Lu, J.; Deng, J.; Wang, C.; He, J.; and Zhang, T. 2023{\natexlab{a}}.
\newblock Query refinement transformer for 3d instance segmentation.
\newblock In \emph{Proceedings of the IEEE/CVF International Conference on Computer Vision}, 18516--18526.

\bibitem[{Lu et~al.(2023{\natexlab{b}})Lu, Xu, Wei, Xie, Tomizuka, Keutzer, and Zhang}]{lu2023ov-3det}
Lu, Y.; Xu, C.; Wei, X.; Xie, X.; Tomizuka, M.; Keutzer, K.; and Zhang, S. 2023{\natexlab{b}}.
\newblock Open-vocabulary point-cloud object detection without 3d annotation.
\newblock In \emph{Proceedings of the IEEE/CVF conference on computer vision and pattern recognition}, 1190--1199.

\bibitem[{Mao et~al.(2022)Mao, Zhang, Jiang, Chang, and Savva}]{mao2022multiscan}
Mao, Y.; Zhang, Y.; Jiang, H.; Chang, A.~X.; and Savva, M. 2022.
\newblock MultiScan: Scalable RGBD scanning for 3D environments with articulated objects.
\newblock In \emph{Advances in Neural Information Processing Systems}.

\bibitem[{Meng et~al.(2023)Meng, Dai, Chen, Zhang, Chen, Liu, Wang, Wu, Yuan, and Jiang}]{meng2023detectionhub}
Meng, L.; Dai, X.; Chen, Y.; Zhang, P.; Chen, D.; Liu, M.; Wang, J.; Wu, Z.; Yuan, L.; and Jiang, Y.-G. 2023.
\newblock Detection hub: Unifying object detection datasets via query adaptation on language embedding.
\newblock In \emph{Proceedings of the IEEE/CVF Conference on Computer Vision and Pattern Recognition}, 11402--11411.

\bibitem[{Misra, Girdhar, and Joulin(2021)}]{misra20213detr}
Misra, I.; Girdhar, R.; and Joulin, A. 2021.
\newblock An End-to-End Transformer Model for 3D Object Detection.
\newblock In \emph{Proceedings of the IEEE/CVF International Conference on Computer Vision}, 2906--2917.

\bibitem[{Nguyen et~al.(2024)Nguyen, Ngo, Kalogerakis, Gan, Tran, Pham, and Nguyen}]{nguyen2024open3dis}
Nguyen, P.; Ngo, T.~D.; Kalogerakis, E.; Gan, C.; Tran, A.; Pham, C.; and Nguyen, K. 2024.
\newblock Open3dis: Open-vocabulary 3d instance segmentation with 2d mask guidance.
\newblock In \emph{Proceedings of the IEEE/CVF Conference on Computer Vision and Pattern Recognition}, 4018--4028.

\bibitem[{Qi et~al.(2019)Qi, Litany, He, and Guibas}]{qi2019votenet}
Qi, C.~R.; Litany, O.; He, K.; and Guibas, L.~J. 2019.
\newblock Deep hough voting for 3d object detection in point clouds.
\newblock In \emph{proceedings of the IEEE/CVF International Conference on Computer Vision}, 9277--9286.

\bibitem[{Rukhovich, Vorontsova, and Konushin(2022)}]{rukhovich2022fcaf3d}
Rukhovich, D.; Vorontsova, A.; and Konushin, A. 2022.
\newblock FCAF3D: fully convolutional anchor-free 3D object detection.
\newblock In \emph{IEEE/CVF European Conference on Computer Vision (ECCV)}, 477--493. Springer.

\bibitem[{Rukhovich, Vorontsova, and Konushin(2023)}]{rukhovich2023tr3d}
Rukhovich, D.; Vorontsova, A.; and Konushin, A. 2023.
\newblock Tr3d: Towards real-time indoor 3d object detection.
\newblock In \emph{2023 IEEE International Conference on Image Processing (ICIP)}, 281--285. IEEE.

\bibitem[{Schult et~al.(2023)Schult, Engelmann, Hermans, Litany, Tang, and Leibe}]{schult2023mask3d}
Schult, J.; Engelmann, F.; Hermans, A.; Litany, O.; Tang, S.; and Leibe, B. 2023.
\newblock Mask3d: Mask transformer for 3d semantic instance segmentation.
\newblock In \emph{2023 IEEE International Conference on Robotics and Automation (ICRA)}, 8216--8223. IEEE.

\bibitem[{Shen et~al.(2023)Shen, Geng, Yuan, Lin, Liu, Wang, Hu, Zheng, and Guo}]{shen2023v-detr}
Shen, Y.; Geng, Z.; Yuan, Y.; Lin, Y.; Liu, Z.; Wang, C.; Hu, H.; Zheng, N.; and Guo, B. 2023.
\newblock V-detr: Detr with vertex relative position encoding for 3d object detection.
\newblock \emph{arXiv preprint arXiv:2308.04409}.

\bibitem[{Shi et~al.(2021)Shi, Zhang, Xu, Dai, Zou, Xiong, and Tian}]{shi2021multi}
Shi, B.; Zhang, X.; Xu, H.; Dai, W.; Zou, J.; Xiong, H.; and Tian, Q. 2021.
\newblock Multi-dataset pretraining: A unified model for semantic segmentation.
\newblock \emph{arXiv preprint arXiv:2106.04121}.

\bibitem[{Song, Lichtenberg, and Xiao(2015)}]{song2015sunrgbd}
Song, S.; Lichtenberg, S.~P.; and Xiao, J. 2015.
\newblock Sun rgb-d: A rgb-d scene understanding benchmark suite.
\newblock In \emph{Proceedings of the IEEE conference on computer vision and pattern recognition}, 567--576.

\bibitem[{Soum-Fontez, Deschaud, and Goulette(2023)}]{soum2023mdt3d}
Soum-Fontez, L.; Deschaud, J.-E.; and Goulette, F. 2023.
\newblock Mdt3d: Multi-dataset training for lidar 3d object detection generalization.
\newblock In \emph{2023 IEEE/RSJ International Conference on Intelligent Robots and Systems (IROS)}, 5765--5772. IEEE.

\bibitem[{Takmaz et~al.(2024)Takmaz, Fedele, Sumner, Pollefeys, Tombari, and Engelmann}]{takmaz2024openmask3d}
Takmaz, A.; Fedele, E.; Sumner, R.; Pollefeys, M.; Tombari, F.; and Engelmann, F. 2024.
\newblock OpenMask3D: Open-Vocabulary 3D Instance Segmentation.
\newblock \emph{Advances in Neural Information Processing Systems}, 36.

\bibitem[{Wald et~al.(2019)Wald, Avatisyan, Navab, Tombari, and Niessner}]{wald20193rscan}
Wald, J.; Avatisyan, A.; Navab, N.; Tombari, F.; and Niessner, M. 2019.
\newblock RIO: 3D Object Instance Re-Localization in Changing Indoor Environments.
\newblock In \emph{Proceedings of IEEE International Conference on Computer Vision (ICCV)}.

\bibitem[{Wang et~al.(2022{\natexlab{a}})Wang, Ding, Dong, Shi, Li, Li, Li, and Wang}]{wang2022cagroup3d}
Wang, H.; Ding, L.; Dong, S.; Shi, S.; Li, A.; Li, J.; Li, Z.; and Wang, L. 2022{\natexlab{a}}.
\newblock Cagroup3d: Class-aware grouping for 3d object detection on point clouds.
\newblock \emph{Advances in Neural Information Processing Systems}, 35: 29975--29988.

\bibitem[{Wang et~al.(2022{\natexlab{b}})Wang, Shi, Yang, Fang, Qian, Li, Schiele, and Wang}]{wang2022rbgnet}
Wang, H.; Shi, S.; Yang, Z.; Fang, R.; Qian, Q.; Li, H.; Schiele, B.; and Wang, L. 2022{\natexlab{b}}.
\newblock RBGNet: Ray-Based Grouping for 3D Object Detection.
\newblock In \emph{IEEE Conference on Computer Vision and Pattern Recognition (CVPR)}, 1110--1119.

\bibitem[{Wang et~al.(2023)Wang, Li, Chen, Lim, Torralba, Zhao, and Wang}]{wang2023unidetector}
Wang, Z.; Li, Y.; Chen, X.; Lim, S.-N.; Torralba, A.; Zhao, H.; and Wang, S. 2023.
\newblock Detecting everything in the open world: Towards universal object detection.
\newblock In \emph{Proceedings of the IEEE/CVF Conference on Computer Vision and Pattern Recognition}, 11433--11443.

\bibitem[{Wang et~al.(2024)Wang, Li, Chen, Zhao, and Wang}]{wang2024uni3detr}
Wang, Z.; Li, Y.-L.; Chen, X.; Zhao, H.; and Wang, S. 2024.
\newblock Uni3detr: Unified 3d detection transformer.
\newblock \emph{Advances in Neural Information Processing Systems}, 36.

\bibitem[{Wu et~al.(2022)Wu, Shi, Du, Lu, Cao, and Zhong}]{wu2022dknet}
Wu, Y.; Shi, M.; Du, S.; Lu, H.; Cao, Z.; and Zhong, W. 2022.
\newblock 3d instances as 1d kernels.
\newblock In \emph{European Conference on Computer Vision}, 235--252. Springer.

\bibitem[{Xie et~al.(2021)Xie, Lai, Wu, Wang, Lu, Wei, and Wang}]{xie2021venet}
Xie, Q.; Lai, Y.-K.; Wu, J.; Wang, Z.; Lu, D.; Wei, M.; and Wang, J. 2021.
\newblock VENet: Voting Enhancement Network for 3D Object Detection.
\newblock In \emph{Proceedings of the IEEE/CVF International Conference on Computer Vision}, 3712--3721.

\bibitem[{Xie et~al.(2020)Xie, Lai, Wu, Wang, Zhang, Xu, and Wang}]{xie2020mlcvnet}
Xie, Q.; Lai, Y.-K.; Wu, J.; Wang, Z.; Zhang, Y.; Xu, K.; and Wang, J. 2020.
\newblock Mlcvnet: Multi-level context votenet for 3d object detection.
\newblock In \emph{Proceedings of the IEEE/CVF conference on computer vision and pattern recognition}, 10447--10456.

\bibitem[{Yeshwanth et~al.(2023)Yeshwanth, Liu, Nie{\ss}ner, and Dai}]{yeshwanthliu2023scannetpp}
Yeshwanth, C.; Liu, Y.-C.; Nie{\ss}ner, M.; and Dai, A. 2023.
\newblock ScanNet++: A High-Fidelity Dataset of 3D Indoor Scenes.
\newblock In \emph{Proceedings of the International Conference on Computer Vision ({ICCV})}.

\bibitem[{Yi et~al.(2019)Yi, Zhao, Wang, Sung, and Guibas}]{yi2019gspn}
Yi, L.; Zhao, W.; Wang, H.; Sung, M.; and Guibas, L.~J. 2019.
\newblock Gspn: Generative shape proposal network for 3d instance segmentation in point cloud.
\newblock In \emph{Proceedings of the IEEE/CVF conference on computer vision and pattern recognition}, 3947--3956.

\bibitem[{Zhang et~al.(2023)Zhang, Yuan, Shi, Chen, Li, and Qiao}]{zhang2023uni3d}
Zhang, B.; Yuan, J.; Shi, B.; Chen, T.; Li, Y.; and Qiao, Y. 2023.
\newblock Uni3d: A unified baseline for multi-dataset 3d object detection.
\newblock In \emph{Proceedings of the IEEE/CVF Conference on Computer Vision and Pattern Recognition}, 9253--9262.

\bibitem[{Zhang et~al.(2024)Zhang, Li, Zhang, Xie, Xue, Xie, and Zhang}]{zhang2024fm-ov3d}
Zhang, D.; Li, C.; Zhang, R.; Xie, S.; Xue, W.; Xie, X.; and Zhang, S. 2024.
\newblock FM-OV3D: Foundation Model-Based Cross-Modal Knowledge Blending for Open-Vocabulary 3D Detection.
\newblock In \emph{Proceedings of the AAAI Conference on Artificial Intelligence}, volume~38, 16723--16731.

\bibitem[{Zhang et~al.(2020)Zhang, Sun, Yang, and Huang}]{zhang2020h3dnet}
Zhang, Z.; Sun, B.; Yang, H.; and Huang, Q. 2020.
\newblock H3dnet: 3d object detection using hybrid geometric primitives.
\newblock In \emph{European Conference on Computer Vision}, 311--329. Springer.

\bibitem[{Zhao et~al.(2020)Zhao, Schulter, Sharma, Tsai, Chandraker, and Wu}]{zhao2020object}
Zhao, X.; Schulter, S.; Sharma, G.; Tsai, Y.-H.; Chandraker, M.; and Wu, Y. 2020.
\newblock Object detection with a unified label space from multiple datasets.
\newblock In \emph{Computer Vision--ECCV 2020: 16th European Conference, Glasgow, UK, August 23--28, 2020, Proceedings, Part XIV 16}, 178--193. Springer.

\bibitem[{Zheng et~al.(2022)Zheng, Duan, Lu, Zhou, and Tian}]{zheng2022hyperdet3d}
Zheng, Y.; Duan, Y.; Lu, J.; Zhou, J.; and Tian, Q. 2022.
\newblock HyperDet3D: Learning a Scene-conditioned 3D Object Detector.
\newblock In \emph{IEEE Conference on Computer Vision and Pattern Recognition (CVPR)}, 5575--5584.

\bibitem[{Zhou, Koltun, and Kr{\"a}henb{\"u}hl(2022)}]{zhou2022simple}
Zhou, X.; Koltun, V.; and Kr{\"a}henb{\"u}hl, P. 2022.
\newblock Simple multi-dataset detection.
\newblock In \emph{Proceedings of the IEEE/CVF conference on computer vision and pattern recognition}, 7571--7580.

\bibitem[{Zhu et~al.(2024)Zhu, Hui, Shen, and Xie}]{zhu2024spgroup3d}
Zhu, Y.; Hui, L.; Shen, Y.; and Xie, J. 2024.
\newblock SPGroup3D: Superpoint Grouping Network for Indoor 3D Object Detection.
\newblock In \emph{Proceedings of the AAAI Conference on Artificial Intelligence}, volume~38, 7811--7819.

\end{thebibliography}

\end{document}